%% file: main.tex
\documentclass[11pt]{article}

\usepackage[final]{acl}

\usepackage{times}
\usepackage{latexsym}
\usepackage{booktabs}
\usepackage{amsmath} 
\usepackage[table]{xcolor}
\usepackage{amssymb} 
\usepackage[most]{tcolorbox} 
\usepackage{cuted}
\usepackage{placeins}
\usepackage{caption}
\usepackage{multirow}
\usepackage{xspace}
\usepackage{float}

\usepackage{graphicx}
\usepackage{fontawesome5}
\usepackage{adjustbox} 
\usepackage{graphicx}

\graphicspath{{figure/}}

\newcommand{\mcpicon}[1]{%
    \raisebox{-0.2\height}{\includegraphics[height=1.2em]{mcp_icon/#1}}%
}

\tcbset{
    desatbox/.style={
        enhanced jigsaw,
        breakable, 
        colback=#1!10!white, 
        colframe=#1!40!gray, 
        coltitle=black!80,
        fonttitle=\bfseries\large,
        boxrule=1pt,
        arc=2mm,
        left=6mm, right=6mm, top=6mm, bottom=6mm,
        parbox=false 
    }
}

\newcommand{\name}[0]{ACE-Router\xspace}
\usepackage[T1]{fontenc}

\usepackage[utf8]{inputenc}

\usepackage{microtype}

\usepackage{inconsolata}

\usepackage{graphicx}

\title{\name: Generalizing History-Aware Routing from MCP Tools to the Agent Web}

\author{
  \textbf{Zhiyuan Yao\textsuperscript{1,}}\thanks{Equal contribution.},
  \textbf{Zishan Xu\textsuperscript{2, *}},
  \textbf{Yifu Guo\textsuperscript{4, *}},
  \textbf{Zhiguang Han\textsuperscript{5}},
  \textbf{Cheng Yang\textsuperscript{6}},
\\
  \textbf{Shuo Zhang},
  \textbf{Weinan Zhang\textsuperscript{2,}}\thanks{Corresponding authors.},
  \textbf{Xingshan Zeng\textsuperscript{3, \dag}},
  \textbf{Weiwen Liu\textsuperscript{2, \dag}}
\\
  \textsuperscript{1}Zhejiang University,
  \textsuperscript{2}Shanghai Jiao Tong University,
  \textsuperscript{3}Huawei Technologies Co. Ltd,
\\
  \textsuperscript{4}Sun Yat-sen University,
  \textsuperscript{5}Nanyang Technological University,
  \textsuperscript{6}Hangzhou Dianzi University
}
\begin{document}
\maketitle

\input{sec/abstract}
\input{sec/intro}

\input{sec/related}
\input{sec/method}
\input{sec/experiment}
\input{sec/conclusion}


\section*{Limitations}



Due to computational resource constraints, we exclusively implemented LoRA fine-tuning on the Qwen3-8B architecture. Nevertheless, we posit that our constructed dataset possesses inherent scalability, suggesting that performance gains could be substantially amplified when applied to larger-scale foundation models.

Furthermore, our current routing mechanism is predominantly trained on tool-use data. While preliminary results indicate that this tool-oriented router generalizes effectively to agent retrieval tasks, we plan to develop specialized routing models explicitly tailored for multi-agent scenarios in future work. Ultimately, we aim to extend this routing training paradigm to encompass universal, massive-scale retrieval requirements, such as long-term memory management.

\section*{Acknowledgments}
The work is supported by National Natural Science Foundation of China (62502310,62322603).

\bibliography{main}

\newpage
\appendix
\input{sec/appendix}

\end{document}

%% file: sec/abstract.tex
\begin{abstract}

With the rise of the Agent Web and Model Context Protocol (MCP), the agent ecosystem is evolving into an open collaborative network, exponentially increasing accessible tools. However, current architectures face severe scalability and generality bottlenecks. To address this, we propose ACE-Router, a pipeline for training history-aware routers to empower precise navigation in large-scale ecosystems. By leveraging a dependency-rich candidate Graph to synthesize multi-turn trajectories, we effectively train routers with dynamic context understanding to create the plug-and-play Light Routing Agent. Experiments on the real-world benchmarks MCP-Universe and MCP-Mark demonstrate superior performance. Notably, ACE-Router exhibits critical properties for the future Agent Web: it not only generalizes to multi-agent collaboration with minimal adaptation but also maintains exceptional robustness against noise and scales effectively to massive candidate spaces. These findings provide a strong empirical foundation for universal orchestration in open-ended ecosystems.
Our code is available at \url{https://github.com/euyis1019/ACE-Router}.
\end{abstract}

%% file: sec/intro.tex
\section{Introduction}

\begin{figure}[t]
    \centering
    \includegraphics[width=1\linewidth]{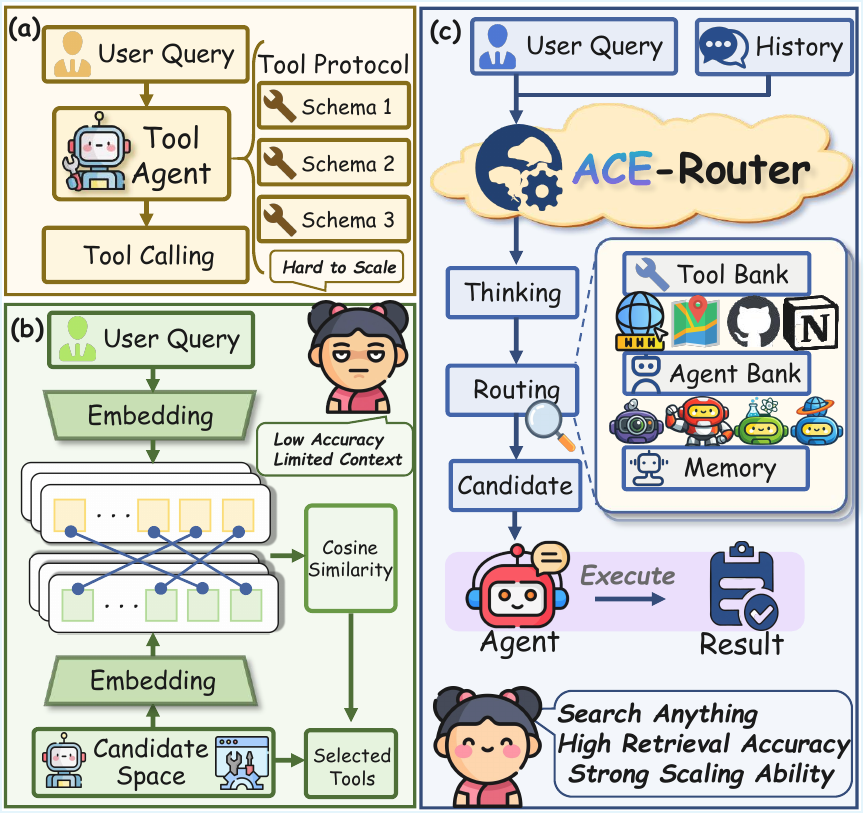}
    \caption{Comparison of \name with other existing paradigms.
(a) \textbf{Static Injection}: Constrained by finite context windows and rigid schemas. 
(b) \textbf{Embedding-based Retrieval}: Limited by static semantic matching and lack of historical context awareness. 
(c) \textbf{\name (Ours)}: A robust router that leverages reasoning and interaction history to achieve high-accuracy retrieval within a massive candidate space.}

    \label{fig:router}
\end{figure}
Recent advances in reasoning and tool utilization have transformed LLMs~\cite{guo2025deepseek,achiam2023gpt} into capable Agents~\cite{tran2025multi,sapkota2025ai,fang2025comprehensive,chang2025grail}. By invoking external tools, they transcend static parametric limits to tackle diverse real-world challenges~\cite{yang2024swe,guo2025octopus,li2025search,xu2026unlockingimplicitexperiencesynthesizing}.  However, most existing systems are monolithic with hardcoded, predefined toolsets, which limit flexibility and prevent seamless integration of different tools and domains. 


To break these boundaries, the emerging Agent Web~\cite{yang2025agentic} envisions an open ecosystem where agents act as autonomous nodes accessing a massive, expanding repository of resources. However, existing multi-agent systems, constrained by static orchestration, are ill-suited for this dynamic scale. To bridge this gap, the paradigm must shift toward "On-demand Teaming": host agents must dynamically discover and schedule optimal collaboration nodes based on real-time states~\cite{lù2025buildwebagentsagents,petrova2025semanticwebmasagentic}. Realizing this adaptive orchestration necessitates a robust \textit{Router}, as illustrated in Figure~\ref{fig:router}(c), capable of navigating the vast search space to identify the most suitable tools ,agents and so on.

Concretely, the Agent Web implies an unprecedented scale of heterogeneous tools and agents, calling for a unified interface to consistently discover and invoke such external resources.
The Model Context Protocol (MCP)~\cite{anthropic2024introducing} is standardizing access to millions of tools. However, current "Static Injection" architectures~\cite{shi2025aimefullyautonomousmultiagentframework,hong2025deepeyesv2, seagent}, as illustrated in Figure~\ref{fig:router} (a), face dual bottlenecks. \textit{Scalability} is restricted by finite context windows, which cannot accommodate massive tool descriptions in a single pass. Meanwhile, \textit{Generality} is undermined by rigid prompt structures, where hard-coded designs lack the flexibility to support dynamic collaboration across heterogeneous architectures.

To manage tool proliferation, retrieval-based tool selection is widely used~\cite{gan2025rag}, yet existing selectors typically rely on static embedding-based matching~\cite{mo2025livemcpbench,qin2023toolllm}, as shown in Figure~\ref{fig:router} (b). However, this approach faces three critical limitations: 
(1) It lacks fine-grained discriminability for functionally similar tools due to semantic overlap; 
(2) It typically ignores the multi-turn trajectory, omitting crucial state information like intermediate outcomes, historical performance, and tool correlations; 
(3) Even if history is incorporated, encoding long contexts into fixed-size vectors causes information compression, failing to resolve subtle distinctions in complex agent states.
Consequently, this precludes the model from leveraging past interactions for informed, context-aware decisions.

To bridge these gaps, we propose \textbf{\name}, a pipeline for training high-performance, history-aware routers. Our approach begins with Graph-based Expansion, which employs self-evolutionary mutation to synthesize behaviorally diverse tools within a structured Candidate Graph, enabling the distinction of subtle functional nuances. Building on this, we implement Trajectory Synthesis by sampling tool subsets via random walks. These subsets drive a multi-agent framework to generate context-rich trajectories, yielding explicit supervision signals that align multi-turn histories with correct routing decisions.
Finally, we introduce the Light Routing Agent, a plug-and-play module that operates through a minimal interface (i.e., Router Invocation and Execution tools). This abstraction decouples routing logic from specific tool definitions, improving generality and enabling seamless adaptation across diverse architectures.


Experimental results demonstrate that \name achieves superior performance on real-world MCP benchmarks. 
Crucially, \name unveils \textit{Cross-domain Transferability}, generalizing to multi-agent tasks with minimal adaptation. Furthermore, it demonstrates \textit{Robustness against Noise}, effectively filtering out irrelevant distractions and hard negatives within massive candidate spaces.

Overall, our contributions are summarized as follows:

\begin{itemize}
    \item We propose \textbf{\name}, a router training framework that integrates graph-based tool expansion and multi-agent trajectory synthesis. By rigorously aligning multi-turn history with routing decisions, this framework constructs high-quality supervision specifically tailored for router training.
    
    \item We train a history-aware router that effectively captures dynamic, multi-turn dependencies. This model transcends the limitations of static semantic matching by maintaining precise context awareness throughout complex interaction trajectories.
    
    \item We develop the Light Routing Agent, a plug-and-play module designed for both tool and agent selection. Experimental results demonstrate its superior performance and robustness on MCP benchmarks and validate its seamless generalization from tool routing to multi-agent orchestration.
\end{itemize}

%% file: sec/related.tex
\section{Related Work}

\subsection{Large-Scale Tool Learning}

With the emergence of open protocols like MCP, the tool ecosystem is transitioning from closed to open systems. This paradigm shift has spurred the development of diverse MCP-specific evaluation benchmarks, ranging from large-scale coverage and multi-domain diversity~\cite{fan2025mcptoolbench,luo2025mcp,mo2025livemcpbench} to real-world service integration~\cite{wu2025mcpmark,guo2025mcpagentbench,mo2025livemcpbench} and multi-dimensional frameworks assessing accuracy, efficiency, and latency~\cite{gao2025mcpradar,luo2025mcp}. However, existing tool learning methods face two fundamental bottlenecks under this new paradigm.

\textbf{Scalability Bottlenecks.} Mainstream approaches adopt two architectural patterns. Hard-coding predefined tool sets into system prompts leads to context saturation as tool numbers grow~\cite{yao2023react,schick2023toolformer,shen2023hugginggpt,yang2024swe}. Alternatively, "retrieve-inject" pipelines filter tool subsets through retrieval before context injection~\cite{patil2023gorilla,qin2023toolllm,zhang2024toolnet,song2023restgpt,scalemcp}, though processing numerous schemas still incurs substantial context overhead.

\textbf{Training Data Gaps.} Current datasets~\cite{li2023apibank,qin2023toolllm} operate at scales below MCP levels. Traditional synthesis methods~\cite{
li2024toolsandbox,yan2024bfcl,chen2024teval} generate isolated query-tool pairs from flat collections, lacking multi-step reasoning patterns and inter-tool dependencies.

\subsection{Dynamic Tool Routing}

\textbf{Static Semantic Matching.} Tool selection typically relies on embedding-based similarity between user queries and tool descriptions~\cite{patil2023gorilla,song2023restgpt}, where single-turn matching determines relevance without considering multi-turn dynamics.

\textbf{Context-Aware Approaches.} Recent methods incorporate execution context through two paradigms: statistics-driven approaches match via probabilistic patterns in usage history~\cite{yang2024autotool,chen2024dtdr}, while graph-based methods model dependencies through neural networks or search algorithms~\cite{zhang2024toolnet,cai2024toolchain,hu2024sitgraph}. These approaches compress dialogue information into fixed representations without directly leveraging raw conversation history for routing decisions in evolving multi-agent scenarios.

\begin{figure*}[t]
    \centering
    \includegraphics[width=0.99\linewidth]{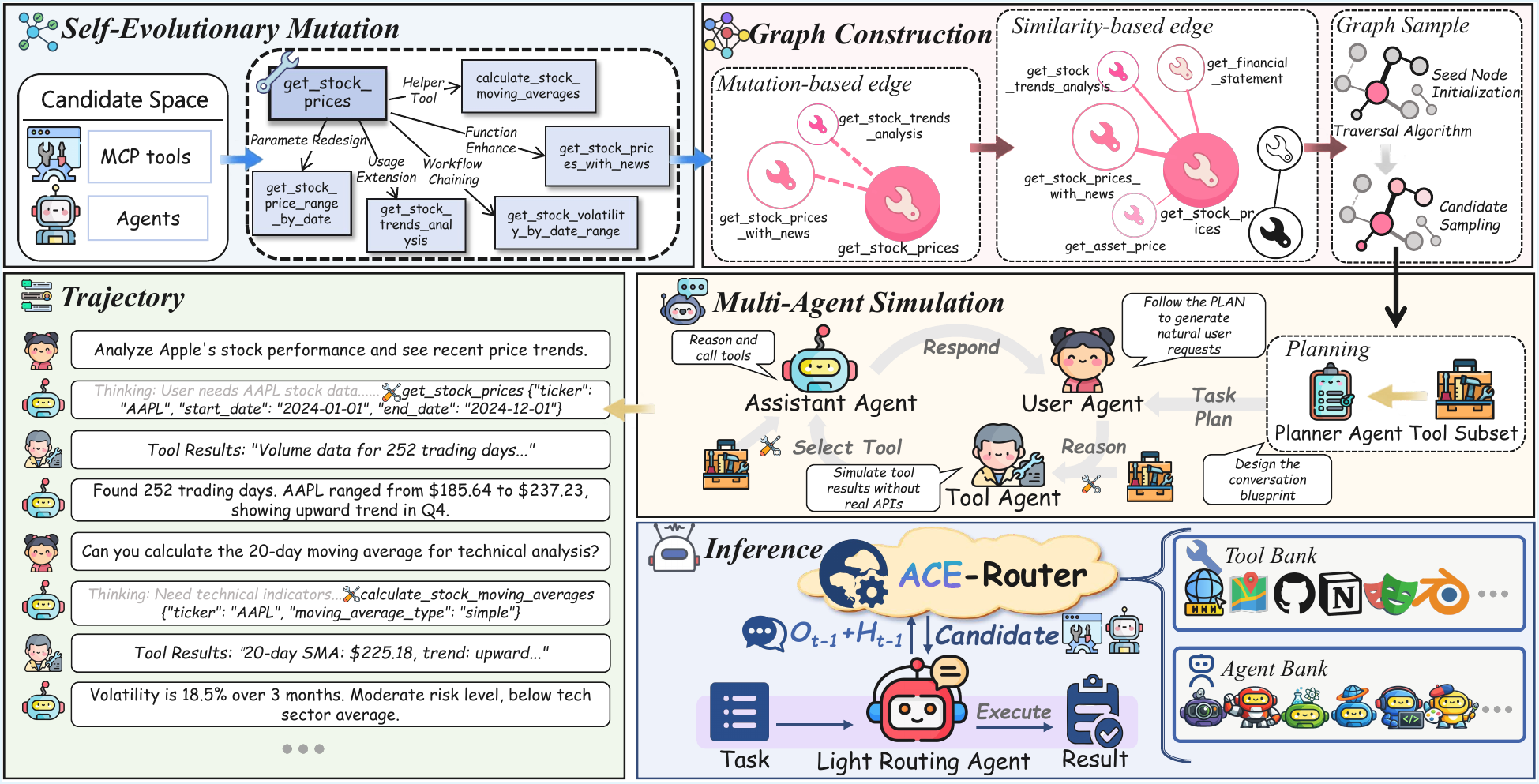}
    \caption{The overall framework of \name. It consists of three key stages: (1) \textbf{Self-evolutionary Graph Construction}, which expands and structures the candidate space via mutation and relation modeling; (2) \textbf{Multi-Agent Simulation}, which synthesizes interaction trajectories to extract history-aware supervision signals; and (3) The \textbf{Light Routing Agent}, designed to seamlessly integrate the trained router into the inference pipeline.}
    \label{fig:framwork}
\end{figure*}

%% file: sec/method.tex
\section{Method}

\subsection{Overview}
Figure~\ref{fig:framwork} illustrates the overall framework of \name. The pipeline operates in a sequential manner: first, we perform a Graph-based Extension with Self-Evolutionary Mutation on the initial candidate set; subsequently, we leverage a multi-agent system to synthesize interaction trajectories, from which we derive supervision signals for the router; finally, we deploy the Light Routing Agent, which serves as the practical implementation of the router trained via \name, designed to seamlessly integrate into existing agent pipelines. We provide a detailed elaboration of these components in the following sections.
\subsection{Problem Formulation.}





We formulate routing as the problem of selecting an appropriate candidate from a given
candidate space $\mathcal{C}$, conditioned on the current user query $Q$ and the dialogue
history $H$.

At each routing step, the candidate space $\mathcal{C}$ is specified beforehand and is
drawn from a predefined set of candidate types, such as a tool set $\mathcal{T}$ or an
agent set $\mathcal{A}$:
\begin{equation}
\mathcal{C} \in \{ \mathcal{T}, \mathcal{A} , ...\}.
\end{equation}

Formally, each candidate $c \in \mathcal{C}$ is associated with a  structured specification
$\phi(c)$.
For tools, $\phi(c)$ includes the tool description and schema, while for agents, $\phi(c)$
corresponds to the agent profile and its available tool, characterizing the agent's specific capability scope.

Given $(Q, H)$ and the specified candidate space $\mathcal{C}$, we train a parameterized
router $\pi_\theta$ to model a conditional distribution over candidates within
$\mathcal{C}$:
\begin{equation}
\pi_\theta(c \mid Q, H, \mathcal{C}), \quad c \in \mathcal{C}.
\end{equation}

At inference time, the router selects the candidate with the highest posterior
probability:
\begin{equation}
c^* = \arg\max_{c \in \mathcal{C}} \pi_\theta(c \mid Q, H, \mathcal{C}).
\end{equation}

\subsection{Candidate Graph-based Extension With Self-Evolutionary Mutation}



The goal of the router is to select candidates that best match the current state from a
set of semantically and functionally related options.
To support this objective during trajectory synthesis, it is crucial to expose the
router to candidates that are not only relevant to the current query, but also closely
related in terms of functionality or dependency structure. 

To effectively scale the candidate space and bolster the discriminative capability against semantically close candidates, we construct a candidate graph over the initial candidate set, where nodes
correspond to candidates (e.g., tools or agents), and edges capture semantic similarity
or functional dependencies between them.
Building on this graph, we further enrich the candidate space through a self-evolutionary mutation process that synthesizes new candidate variants from existing ones, inspired by methods such as Self-Evolve~\cite{novikov2025alphaevolve}.

\subsubsection{Graph Construction}





Given a candidate set $\mathcal{C} = \{ c_1, c_2, \ldots, c_N \}$, we first derive a vector representation for each candidate by encoding its structured specification $\phi(c)$. Using a pretrained embedding model $\mathcal{E}$, the embedding vector $\mathbf{h}_i \in \mathcal{R}^d$ for candidate $c_i$ is computed as $\mathbf{h}_i = \mathcal{E}(\phi(c_i))$. For instance, within the tool subset $\mathcal{T} \subseteq \mathcal{C}$, $\phi(c_i)$ serializes the tool's textual description and input schema.

To capture latent relationships, we define the semantic similarity between any pair of candidates $c_i$ and $c_j$ as the cosine similarity of their embeddings:
\begin{equation}
    \mathrm{sim}(c_i, c_j) = \cos(\mathbf{h}_i, \mathbf{h}_j) = \frac{\mathbf{h}_i \cdot \mathbf{h}_j}{\| \mathbf{h}_i \| \| \mathbf{h}_j \|}.
\end{equation}
We construct an undirected edge between nodes $c_i$ and $c_j$ if their similarity exceeds a predefined threshold $\tau$ (empirically set to $0.82$), i.e., $\mathrm{sim}(c_i, c_j) > \tau$. This procedure yields an initial connectivity graph $\mathcal{G} = (\mathcal{C}, \mathcal{E}_{\text{sim}})$, which captures the local semantic neighborhoods among candidates.

\subsubsection{Self-Evolutionary Mutation}


To mitigate overfitting caused by an overly narrow candidate space $\mathcal{C}$, we
introduce a novel Self-Evolutionary strategy to construct new candidate elements.
The key idea is to iteratively expand the candidate graph with controlled mutations
that preserve semantic relevance to existing candidates.

Specifically, we define a set of mutation operators $\mathcal{M}$ for tool, which include
\emph{Function Enhancement}, \emph{Parameter Mutation}, \emph{Workflow Chaining},
\emph{Helper Operation}, and \emph{Usage Extension}. Detailed specifications of these operators, along with the agent mutation strategies, are provided in Appendix~\ref{app:mutation}.
At each iteration, we randomly sample an existing candidate $c \in \mathcal{C}$ and
a mutation operator $m \in \mathcal{M}$.
We then prompt a large language model (LLM) to synthesize a new candidate
$c' = m(c)$ based on the selected mutation.

The newly generated candidate $c'$ is added as a new node to the candidate graph, and
an edge is created between $c'$ and the original candidate $c$ to explicitly encode
their mutation relationship.
This Self-Evolutionary process progressively enriches the candidate space while maintaining
local structural consistency in the graph.

\subsection{History-Aware Supervision for Router Training}

We begin by sampling candidate subsets from the constructed candidate graph via a random walk--based traversal, aiming to select candidates that exhibit semantic similarity or functional dependencies. Specifically, we initiate the process from a set of seed nodes and perform a DFS-style traversal to visit neighboring nodes. The collected nodes form a sampled subset, ensuring local coherence in terms of semantics and functionality.

Inspired by prior work on tool-oriented trajectory synthesis~\cite{liu2024toolace,wang2025toolflow}, we further synthesize a task description and a coarse-grained execution plan conditioned on the sampled subset. Based on this plan, we generate multi-turn dialogue trajectories through role-based simulation. Formally, each trajectory is represented as a sequence:
\begin{equation}
\mathcal{P} = (o_0, a_0, o_1, a_1, \ldots, o_n, a_n),
\end{equation}
where $o_0$ denotes the initial user query, and $o_t$ represents the user feedback or environment response following the assistant action $a_t$ (including invocation results of candidate elements). Crucially, both the trajectory generation and the simulated responses are produced by Large Language Models. This environment-free simulation design enables scalable and flexible synthesis without requiring access to real execution APIs, thereby facilitating the efficient expansion of training data.


Upon acquiring the synthesized trajectories, we proceed to extract supervision signals for router training. Specifically, we identify time steps $t$ where the assistant action $a_t$ involves invoking a specific candidate $c \in \mathcal{C}$, which we extract as the ground-truth label. To construct the corresponding input, we designate the preceding interaction sequence $(o_0, \dots, a_{t-1})$ as the history context $H$, while explicitly treating the immediate observation $o_{t-1}$ as the current query. This strategy effectively transforms complex, multi-step trajectories into a large-scale dataset of history-aware routing instances. Depending on the definition of the candidate space $\mathcal{C}$, these supervision signals can be universally applied to train various router types, including both Tool Routers and Agent Routers. As validated in Section~\ref{exp:Significance_history}, this history-aware formulation yields significant accuracy gains over stateless baselines.

\subsection{Light Routing Agent}
\input{table/universe_and_mark}



To seamlessly integrate the trained router into existing agent workflows and evaluation benchmarks,
we design a lightweight routing agent, termed the Light Routing Agent (LRA),
which decouples routing decisions from concrete task execution.
Unlike conventional agents that tightly couple planning, tool selection, and execution logic,
LRA serves solely as a minimal wrapper around the trained router.

Specifically, LRA is equipped with only two tools.
The first is a router invocation tool, which queries the trained router based on the current dialogue history and contextual information
to select the most appropriate candidate from a given candidate set.
The second is an execution tool, responsible for invoking or executing the candidate returned by the router.
With this design, the agent no longer needs to explicitly inject large candidate set information
(e.g., tool descriptions or agent functionalities) into the context.
Instead, it dynamically selects and dispatches the required operations at runtime via the router,
thereby maintaining a lightweight agent structure while enabling efficient execution of diverse and complex tasks.

%% file: table/universe_and_mark.tex
\begin{table*}[t]
\centering
\setlength{\tabcolsep}{3pt} 
\caption{
    \textbf{Accuracy (\%) comparison on MCP-Universe and MCP-Mark benchmarks.}
    \textbf{Q} denotes methods using only the current query, while \textbf{Q+H} incorporates both the query and interaction history.
    For \textbf{MCP-Universe}, we evaluate six specific domains: Location Navigation (Loc.), Repository Management (Repo.), Financial Analysis (Fin.), 3D Designing (3D), Browser Automation (Browser), and Web Searching (Web).
    \textbf{MCP-Mark} assesses performance on specific real-world tool environments including Notion, GitHub, PostgreSQL, Playwright, and Filesystem. The best result is marked in \textbf{bold} and the second best result is \underline{underlined}. 
}
\label{tab:mcp_full}
\resizebox{\textwidth}{!}{
\begin{tabular}{lccccccccccccc}
\toprule
\textbf{Methods} &
\multicolumn{7}{c}{\textbf{MCP-Universe}} &
\multicolumn{6}{c}{\textbf{MCP-Mark}} \\
\cmidrule(lr){2-8} \cmidrule(lr){9-14}
 &
\mcpicon{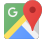}Loc. & \mcpicon{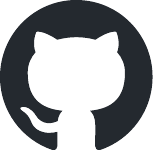}Repo. & \mcpicon{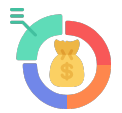}Fin. & \mcpicon{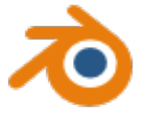}3D & \mcpicon{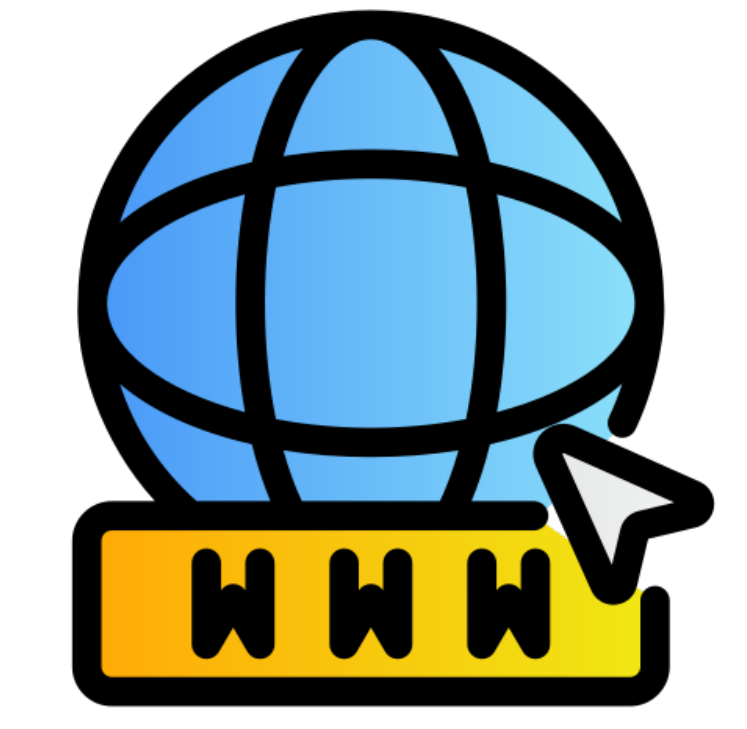}Browser & \mcpicon{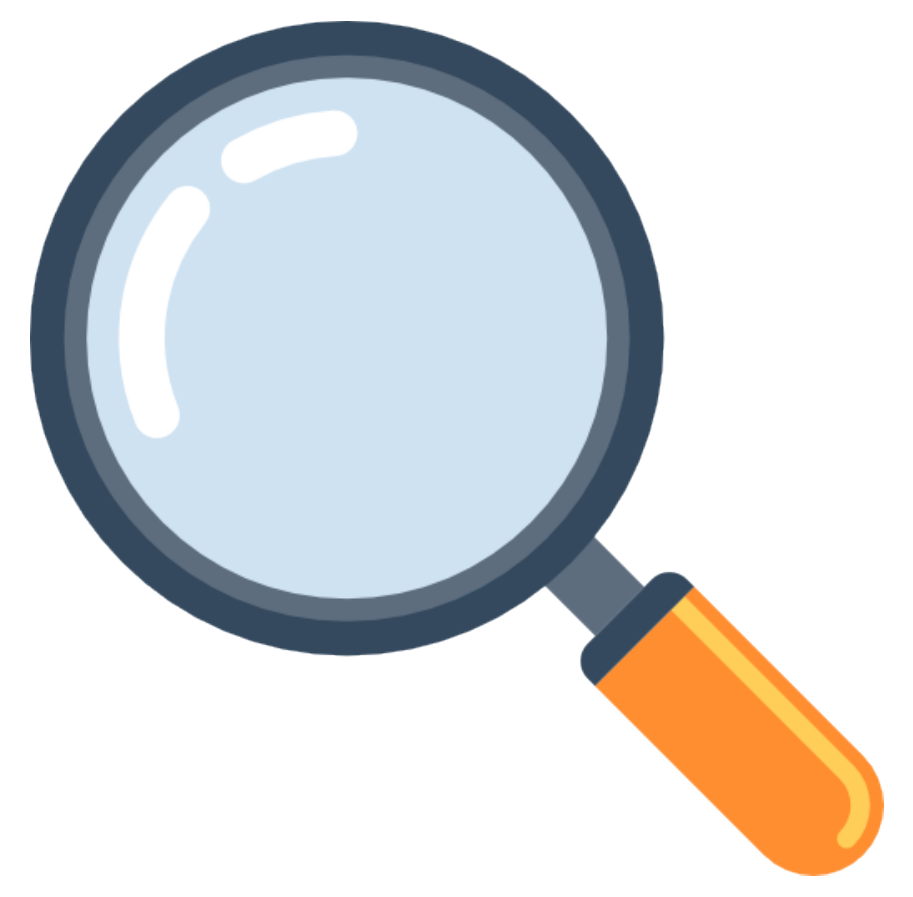}Web & Overall &
\mcpicon{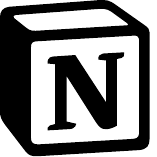}Notion & \mcpicon{github}GitHub & \mcpicon{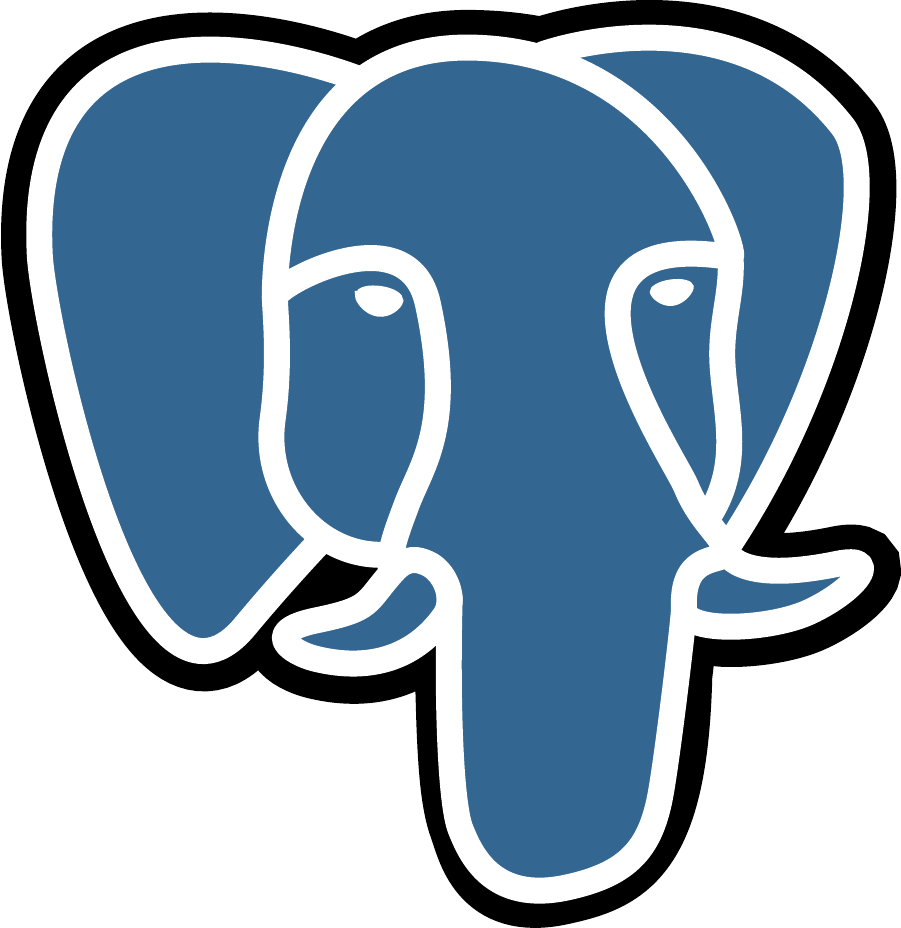}PostgreSQL & \mcpicon{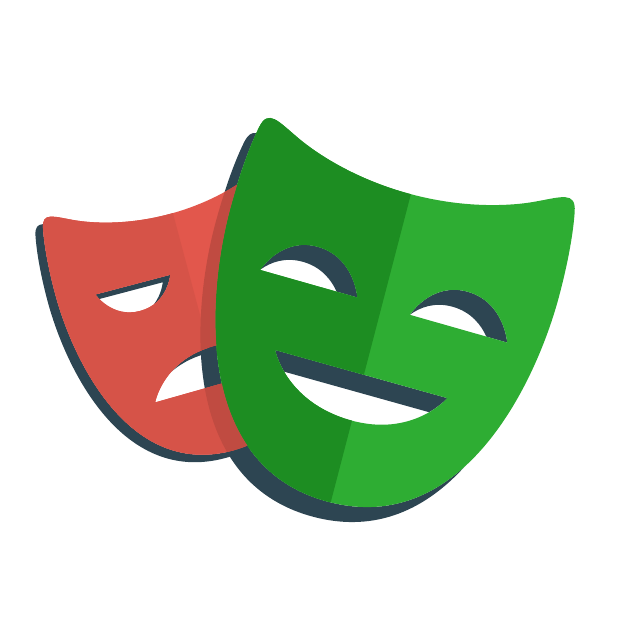}Playwright & \mcpicon{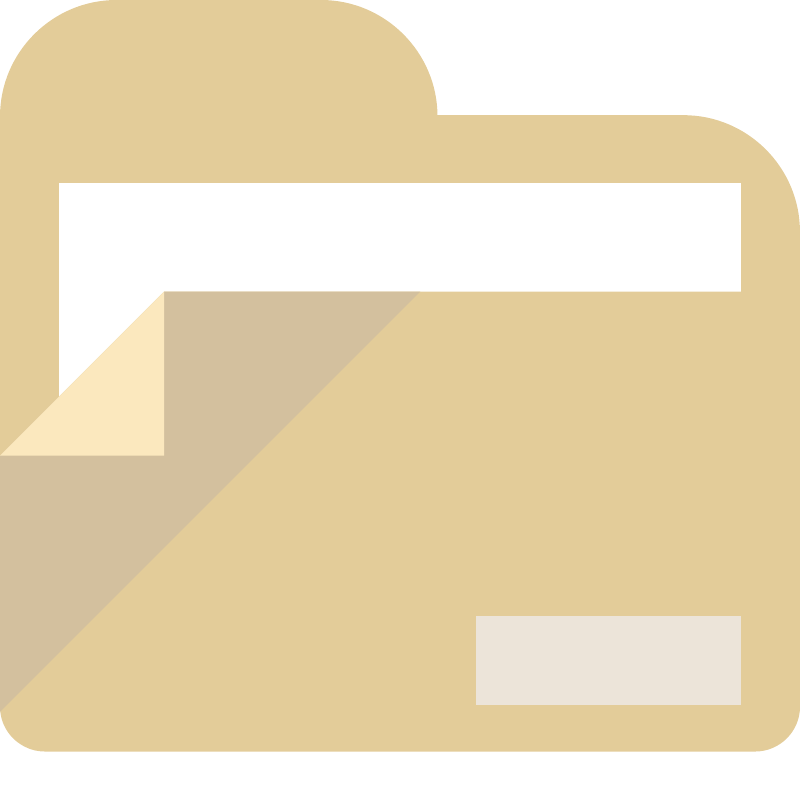}Filesystem & Overall \\
\midrule

\rowcolor{gray!12}
\multicolumn{14}{c}{\textit{Embedding-based Retrieval}}\\
Text-Emb-3-Large  (Q)    & 46.11 & 38.62 &62.71  & 31.58 &28.74 & 16.00 &40.95 & 20.00  &10.00 &70.00  &10.00  & 20.00 & 26.00   \\
Text-Emb-3-Large  (Q+H)  & 45.81 & 14.48 &50.85  & 26.32 & 34.48 &  14.00& 35.20 & 20.00& 0.00 & 60.00 & 10.00 & 10.00 & 20.00     \\
all-MiniLM-L6-v2 (Q)     & 44.91 & 34.48 &  67.80& 26.32 &34.48  & \underline{20.00} &  40.67& 10.00 & 10.00 & 70.00 & 10.00& 30.00 &  26.00   \\
all-MiniLM-L6-v2 (Q+H)   & 52.10 & 33.10  & 66.10 &28.95  & 37.93 & 12.00 & 43.62 &  20.00& 10.00 &80.00  &20.00 & 30.00 &     32.00 \\
\midrule

\rowcolor{gray!12}
\multicolumn{14}{c}{\textit{ReAct Agents}}\\
ReAct (Gemini-2.5-Pro)   & 42.81 & 43.45& 66.10 & 55.26 & 27.59 & 16.00  & 41.80 & 20.00 & 20.00 & \underline{90.00} & 20.00 & 50.00 &  40.00\\
\midrule

\rowcolor{gray!12}
\multicolumn{14}{c}{\textit{LLM-based Routers}}\\
Qwen3-8B          & 48.50 & 49.66 & \underline{69.49} & 50.00 & 29.89 & 18.00 & 46.14 & \underline{30.00} & 20.00 & \underline{90.00} & 20.00 & 50.00 & 42.00 \\
GLM-4.5         & 53.29 & 46.90 & 52.54 & 47.37& 35.63 &  10.00& 46.42  & 20.00 &  10.00& 80.00  & \underline{30.00}& 50.00 & 38.00 \\
DeepSeek-V3.2     & 49.10 & 44.83 & 64.41 & 50.00 &33.33  & 8.00  &44.74  & 20.00 & 10.00 & \textbf{100.00}  & \underline{30.00} & 50.00 &  42.00\\
Claude-sonnet-4 &53.59  & 43.45 & 61.02 & 55.26 & 42.53& 16.00 & 48.25 & \underline{30.00} &\underline{40.00}  &  \underline{90.00}& 20.00 &40.00  &44.00  \\
Gemini-2.5-Pro    &\underline{54.19}  & 46.21 & 61.02 & 55.26 &  \underline{47.13}&  18.00& \underline{49.79} &20.00  & 30.00& \textbf{100.00}  &20.00 & \underline{60.00}  &46.00  \\
Gemini-2.5-flash  &50.90  & \underline{50.34} & 66.10 &52.63  &28.74 &16.00  & 46.98 & \underline{30.00}&\textbf{50.00}&  90.00 & 20.00 & 50.00 & \underline{48.00} \\
GPT-4.1           &45.51  &  48.97&  61.02&  52.63&  34.48&  18.00& 44.60 & 10.00 &20.00  &  90.00& 10.00 & 50.00 &36.00  \\
GPT-4o            & 47.90& 48.97 & 62.71 & \underline{65.79} & 44.83 & 12.00 &47.41  & 20.00 &30.00  & 90.00 & 10.00 &40.00  & 38.00 \\
\rowcolor{blue!20}
\textbf{\name (Ours)}& \textbf{54.49} & \textbf{51.03}  & \textbf{72.88} & \textbf{71.05} &\textbf{49.43}  & \textbf{24.00} & \textbf{53.44} & \textbf{40.00} &\textbf{50.00}  & \textbf{100.00} &\textbf{40.00}  &\textbf{70.00}   & \textbf{60.00} \\
\midrule
\end{tabular}
}
\end{table*}

%% file: sec/experiment.tex
\section{Experiment}
\subsection{Experiment Setup}
\subsubsection{Dataset and model}



Our initial tool bank consisted of 627 MCP tools collected from the MCP Universe~\cite{luo2025mcp} and the LiveMCP~\cite{mo2025livemcpbench} benchmark. By applying the mutation operators described in the previous section, we expanded this initial set into 2,005 tools. For the toolgraph construction, we employed all-MiniLM-L6-v2 to generate semantic embeddings. Subsequently, leveraging GPT-4o for trajectory synthesis, we utilized this augmented tool bank to construct a comprehensive dataset of over 15,092 training samples for the tool router. Although trained on tool selection, the router captures transferable decision patterns, enabling generalization to agent routing tasks without additional training.

Our Tool Router is trained on top of Qwen3-8B~\cite{yang2025qwen3}.
We evaluate the proposed router against a diverse set of baseline methods,
including the native Qwen3-8B model as well as several representative closed-source large language models,
such as GPT-4o~\cite{hurst2024gpt}, Claude-Sonnet-4~\cite{anthropic2025claude4}, Gemini-2.5-Pro~\cite{comanici2025gemini} and so on.
In addition to model-based routing approaches,
we also include embedding-based routing strategies as baseline methods, utilizing all-MiniLM-L6-v2 and text-embedding-3-large as the underlying encoders.
These approaches select candidates by computing vector similarity between the query and candidates,
and we consider multiple input settings,
including using only the current query (query-only),
incorporating historical context. We also include an LLM-driven ReAct~\cite{yao2023react} agent as a baseline for tool selection.

To ensure a fair comparison of routing capability across different models, we fix the downstream execution (reasoning) model to Gemini-2.5-Pro for all router-based methods.

\subsubsection{Benchmark and Evaluation}




We conduct a systematic evaluation of the proposed router on several widely used MCP benchmarks, including MCP-Universe~\cite{luo2025mcp} and MCP-Mark (easy mode)~\cite{wu2025mcpmark}.



To further evaluate the router’s generalization ability in cross-agent scenarios, we construct an evaluation setup tailored to the agent routing task. We systematically collect and normalize over 40 mainstream agents, unifying them into a consistent JSON format to form an initial Agent Bank as the candidate space.Based on the proposed Self-Evolutionary mutation mechanism and multi-agent trajectory synthesis strategy, we generate a total of 156 agent router test samples. All samples are utilized to assess the router’s generalization performance under unseen agent combinations and complex candidate spaces. For a detailed taxonomy of agent Self-Evolutionary mutation types and examples of the benchmark tasks, please refer to Appendix~\ref{app:mutation} and Appendix~\ref{app:route_bench}.

\subsubsection{Implementation Details}

Given resource constraints, We fine-tune the model using LoRA~\cite{hu2022lora} applied to all linear layers with a rank of $r=8$. Training runs for 3 epochs with a global batch size of 64, utilizing a learning rate of $1 \times 10^{-4}$ with a cosine annealing schedule and a 0.1 warmup ratio. The maximum sequence length is set to 32,768 tokens in BF16 precision. For evaluation, we set the sampling temperature to 1 and report the average results over 5 independent runs (avg@5) to ensure stability.

\subsection{Main Result}

As shown in Table~\ref{tab:mcp_full}, \name consistently outperforms all baseline methods, significantly enhancing the agent's capability to solve tasks using MCP tools. Specifically, on the MCP-Universe benchmark, we achieve an overall performance of 53.44\%, with the Financial Analysis domain reaching 72.88\%. On MCP-Mark, our method attains 60.00\%.

Crucially, our results demonstrate that the router-based paradigm significantly surpasses both Embedding-based retrieval and ReAct-based agents. Furthermore, a key finding is that our 8B-parameter specialized router outperforms massive generalist models, including GPT-4o (47.41\% on MCP-Universe) and Gemini-2.5-Pro (49.79\% on MCP-Universe). This highlights a critical limitation in generalist LLMs: despite their reasoning prowess, they struggle with the precise discrimination required for tool selection. Collectively, these findings validate the effectiveness of the \textit{light routing agent} design. Our results demonstrate that employing a specialized router represents a superior strategy for enabling efficient and reliable tool usage, thereby providing a robust foundation for the emerging open Agent Web.




\subsection{Scalability and Robustness Analysis}
\input{table/noise}

To validate the router's adaptability to realistic and challenging scenarios, we evaluate its performance under two distinct conditions: expanded tool spaces and noisy input environments.

\paragraph{Scalability to Large-Scale Tool Spaces.}
We first evaluate the router's scalability by aggregating tools from all MCP servers into a unified candidate pool, extending beyond single-server experiments. This setup allows us to assess performance within a heterogeneous, large-scale tool space. As shown in Table~\ref{tab:robustness_noise}, competing methods experience notable performance degradation when facing this expanded search space. For instance, on the MCP-Universe benchmark, ReAct agents drop from 41.80\% to 36.47\%. In contrast, \name demonstrates exceptional stability, maintaining an accuracy of 53.02\% (marginally shifted from 53.44\%). This demonstrates the effectiveness of \name in handling large candidate pools, suggesting that with increased training data and model capacity, it can reliably scale to retrieve tools from web-scale open tool ecosystems.

\paragraph{Robustness against Tool Noise.}
We evaluate the robustness of the router by introducing additional noisy tools from two distinct sources: (1) \textit{External Benchmarks} (+LiveMCP), which consists of callable tools drawn from real-world environments within the same or related domains; and (2) \textit{Self-Evolutionary Mutations} (+Mutation), comprising automatically synthesized non-callable variants that are semantically similar to the target tools and introduce complex functional dependencies.

Table~\ref{tab:robustness_noise} illustrates the impact of these noisy settings on model performance. On MCP-Mark under the +LiveMCP setting, even advanced generalist models struggle to filter out high-interference noise; for instance, GPT-4o and Gemini-2.5-Pro achieve accuracies of only 28.00\% and 32.00\%, respectively. In contrast, \name demonstrates superior resilience, maintaining a high accuracy of 56.00\%. Similarly, under the +Mutation setting—where injected tools are highly confusable with targets—\name exhibits minimal performance degradation. Specifically, accuracy dips only slightly from 53.44\% to 53.02\% on MCP-Universe and from 60.00\% to 54.00\% on MCP-Mark.


These results indicate that while generalist models are susceptible to distraction, our specialized router remains robust against both real-world noise and fine-grained hard negatives. This resilience stems from our rigorous training methodology, where the self-evolutionary mutation mechanism forces the router to distinguish between targets and semantically close distractors, thereby establishing stable and fine-grained discriminative boundaries.

\subsection{Generalization to Agent Routing}

\begin{figure}[t]
    \centering
    \includegraphics[width=\linewidth]{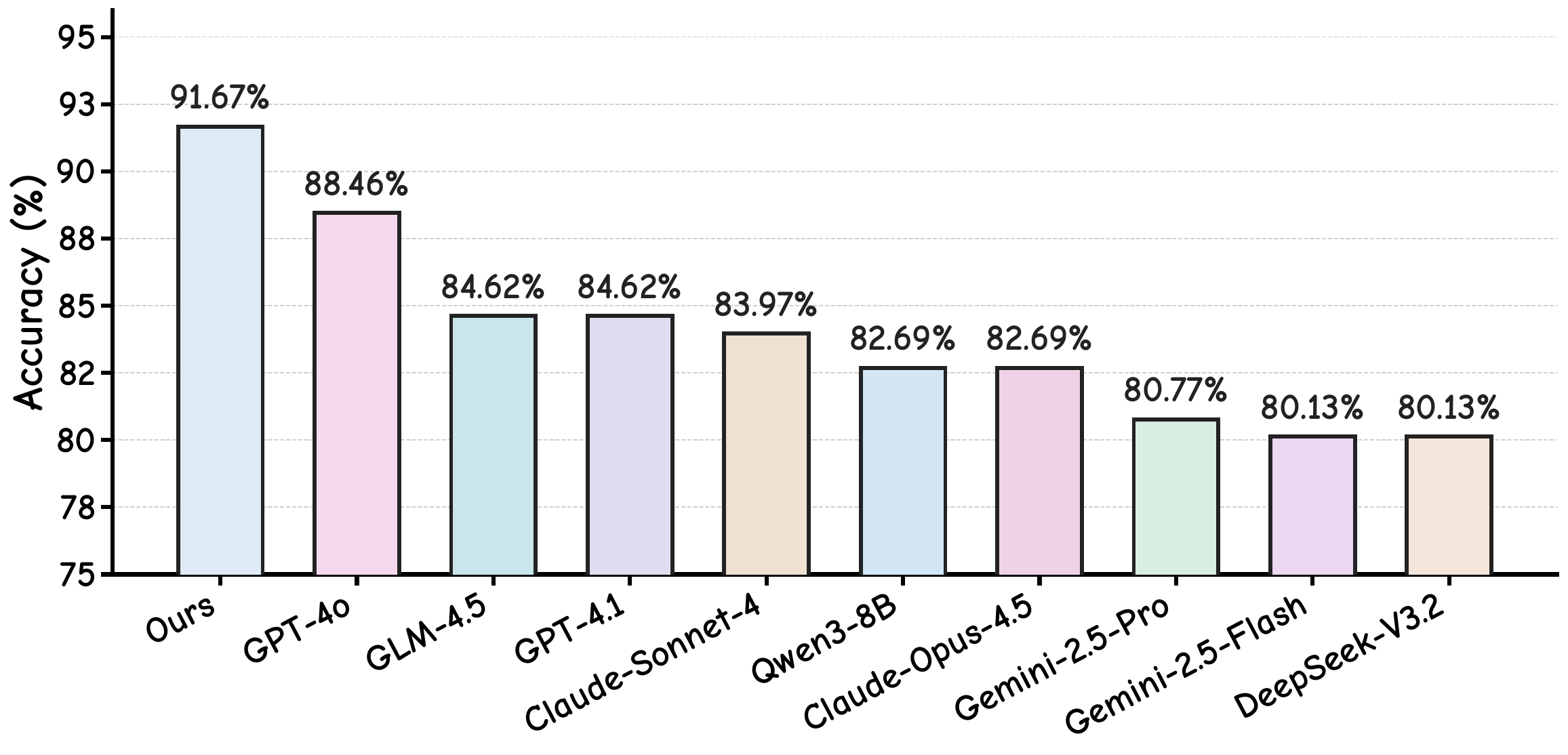}
    \caption{
        \textbf{Performance evaluation on the Agent Route Benchmark.} 
        Comparative analysis of agent route accuracy between \name and representative baselines.
    }
    \label{fig:agent_bench}
\end{figure}


We extended our evaluation to the constructed\textbf{ Agent Route Benchmark}, assessing \name alongside a series of representative state-of-the-art models on their ability to accurately select the optimal agent for subsequent operations based on the given task query and interaction history.

As illustrated in Figure~\ref{fig:agent_bench}, \name significantly outperforms all baselines in agent selection tasks, achieving an accuracy of 91.6\%. This exceptional performance highlights a critical advantage of our approach: it learns the fundamental logic of "capability matching" rather than overfitting to specific tool schemas. Despite being trained primarily on tool data, the router successfully transfers this abstract decision-making pattern to the agent domain without additional fine-tuning. This generalization capability is pivotal for the envisioning of the Agent Web—an interconnected ecosystem comprising millions of specialized agents. In such a decentralized landscape, our router serves as a universal dispatcher, enabling dynamic, on-demand teaming by accurately identifying and orchestrating diverse agents to collaborate on complex tasks, thereby serving as a foundational infrastructure for future multi-agent systems.

\subsection{Significance of History-Aware Routing}
\label{exp:Significance_history}
\begin{figure}[t]
    \centering
    \includegraphics[width=0.99\linewidth]{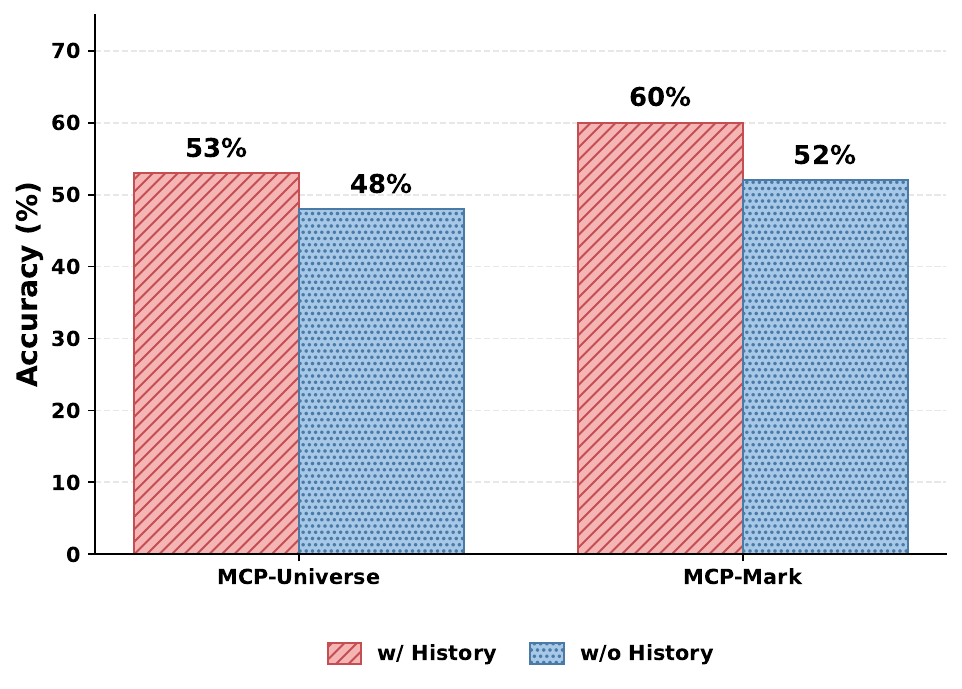} 
    \caption{\textbf{Impact of Historical Context.} A performance comparison between the history-aware model and an ablation variant trained without historical context.}
    \label{fig:history_ablation}
\end{figure}


A distinct advantage of \name lies in its capability to effectively leverage the interaction history of the primary reasoning agent to inform routing decisions. This historical context encapsulates critical information—including intermediate outcomes, prior successes and failures, and latent tool usage correlations—that cannot be adequately captured by the current query alone.

To validate the efficacy of this history-aware mechanism, we conducted an ablation study by intentionally stripping historical context information from the training data. As illustrated in Figure~\ref{fig:history_ablation}, the removal of historical context leads to a significant decline in overall routing accuracy. Specifically, the performance drops from 53\% to 48\% on MCP-Universe and from 60\% to 52\% on MCP-Mark. This significant performance gap underscores the critical role of interaction history in two key dimensions: (1) \textit{Sequential Dependency Reasoning}, where the model must track task progress to respect logical prerequisites—for example, ensuring a user profile is retrieved before attempting to access specific GitHub repository details; and (2) \textit{Error Recovery}, where the model utilizes prior execution feedback to recognize failures and pivot to alternative strategies rather than repeating erroneous calls. These findings confirm that effective routing is inherently a dynamic, history-dependent reasoning process, far exceeding the capabilities of static semantic matching.

\subsection{Ablation on Mutation and Vanilla Fine-Tuning}
\label{exp:mutation_and_vanilla_ablation}

To isolate the contributions of our core components and verify that our performance gains stem from architectural design rather than merely large-scale supervision, we conduct two key ablation studies. First, we evaluate a variant without Self-Evolutionary Mutation (\textit{w/o Mutation}) to demonstrate the critical impact of our dynamic data synthesis. Second, to further isolate the effectiveness of our architectural components, we provide an additional variant (\textit{Vanilla Fine-Tuning}) that removes both the historical context and the Self-Evolutionary Mutation techniques proposed in our paper, relying solely on naive fine-tuning. 

As shown in Table~\ref{tab:comprehensive_ablation}, removing the mutation module causes a substantial performance drop. Expanding the tool candidate pool from 627 to 2,005 effectively mitigates the risk of overfitting to a narrow candidate space. Furthermore, the full \name framework significantly outperforms this vanilla baseline on both benchmarks. This confirms that our core designs---history-aware reasoning and self-evolutionary data synthesis---fundamentally drive the model's strong generalization, providing benefits far beyond standard fine-tuning.

\begin{table}[ht]
    \centering
    \caption{\textbf{Ablation Results.} Performance impact of the Self-Evolutionary Mutation module and comparison with standard fine-tuning baselines.}
    \label{tab:comprehensive_ablation}
    \resizebox{\columnwidth}{!}{%
    \begin{tabular}{lcc}
        \toprule
        \textbf{Method} & \textbf{MCP-Universe (Acc. \%)} & \textbf{MCP-Mark (Acc. \%)} \\
        \midrule
        \textbf{\name (Ours)} & \textbf{53.44} & \textbf{60.00} \\
        w/o Mutation          & 47.12          & 44.00 \\
        Vanilla Fine-Tuning   & 46.70          & 42.00 \\
        \bottomrule
    \end{tabular}%
    }
\end{table}

%% file: table/noise.tex
\begin{table*}[t]
\centering
\small
\setlength{\tabcolsep}{3.5pt} 
\caption{\textbf{Robustness and Scalability Analysis.} We evaluate model performance on MCP-Universe and MCP-Mark under four settings: \textbf{Clean} (single-server baseline), \textbf{Multi} (merged multi-server tools setting), \textbf{+Mutation} (adding Self-Evolutionary mutation tools), and \textbf{+LiveMCP} (adding  real external tools). The best result is marked in \textbf{bold} and the second best result is \underline{underlined}.}
\label{tab:robustness_noise}
\begin{tabular}{l cccc cccc}
\toprule
\multirow{2}{*}{\textbf{Method}} &
\multicolumn{4}{c}{\textbf{MCP-Universe}} &
\multicolumn{4}{c}{\textbf{MCP-Mark}} \\
\cmidrule(lr){2-5} \cmidrule(lr){6-9}
 & Clean & Multi & +Mutation & +LiveMCP & Clean & Multi & +Mutation & +LiveMCP \\
\midrule

\rowcolor{gray!12}
\multicolumn{9}{c}{\textit{Embedding-based Retrieval}}\\
Text-Emb-3-Large (Q)       &40.95  & 40.11 & 39.69 & 39.00 &  26.00& 18.00 & 20.00 & 14.00 \\
Text-Emb-3-Large (Q+H)     &35.20 & 34.23 & 34.37 & 33.24 & 20.00 &14.00  &12.00  & 10.00 \\
all-MiniLM-L6-v2 (Q)       &40.67 & 39.69 & 39.83 & 38.57 & 26.00 &20.00  &22.00  &  14.00\\
all-MiniLM-L6-v2 (Q+H)     &43.62 & 42.92 & 42.78 & 41.79 & 32.00 &26.00  & 20.00 & 18.00 \\
\midrule

\rowcolor{gray!12}
\multicolumn{9}{c}{\textit{ReAct Agents}}\\
ReAct (Gemini-2.5-Pro)     &41.80 & 36.47 & 40.95 & 40.11 &  40.00&32.00  & 26.00 & 20.00 \\
\midrule

\rowcolor{gray!12}
\multicolumn{9}{c}{\textit{LLM-based Routers}}\\
Qwen3-8B                    &46.14 & 45.30 & 44.88 & 44.46& 42.00 &38.00  & 32.00 &30.00  \\
GLM-4.5                    &46.42 & 45.44 & 45.30 & 44.32 & 38.00 & 32.00 & 30.00 &26.00  \\
DeepSeek-V3.2              & 44.74 & 43.76 & 44.04 & 43.07 & 42.00 & 38.00 & 36.00 & 30.00 \\
Claude-Sonnet-4          & 48.25 & 47.41 & 47.13 & 46.84  & 44.00 & \underline{40.00} & 32.00 & 30.00 \\
Gemini-2.5-Pro             & \underline{49.79} &\underline{49.09} & \underline{48.81} & \underline{48.11} & 46.00 & \underline{40.00} & 36.00 & 32.00 \\
Gemini-2.5-flash           & 46.98 & 46.28 & 45.99 & 43.76 & \underline{48.00} &  \underline{40.00}& \underline{44.00} &\underline{36.00}  \\
GPT-4.1                    & 44.60 & 43.90 & 43.48 & 42.64 &  36.00 & 32.00 & 30.00 &28.00  \\
GPT-4o                     & 47.41 & 46.56 & 46.42 & 45.86 & 38.00 & 34.00 & 32.00 & 28.00 \\
\rowcolor{blue!20}
\textbf{\name (Ours)}              & \textbf{53.44} & \textbf{53.02} & \textbf{53.02} & \textbf{52.60} &\textbf{60.00}  & \textbf{56.00} & \textbf{54.00} &\textbf{56.00}  \\
\bottomrule
\end{tabular}
\end{table*}

%% file: sec/conclusion.tex
\section{Conclusion}



In this paper, we introduced \name, a general framework designed for training robust history-aware router models. Our approach begins by expanding an initial candidate pool via self-evolving mutation operators to construct a comprehensive Candidate Graph. Subsequently, we generate effective supervisory signals for the router by employing random walk sampling on the graph coupled with multi-agent trajectory synthesis. Experimental results demonstrate that the router trained on our synthesized data not only achieves superior performance and robustness on MCP tool benchmarks but also exhibits strong generalization capabilities in agent retrieval tasks. These findings pave the way for a router-centric paradigm in future multi-agent collaboration within the Agent Web ecosystem.

%% file: sec/appendix.tex
\section{Comparison with Stronger Retrieval Baselines}
\label{sec:appendix_stronger_baselines}

We expand our evaluation by replacing the initial retrieval baselines with stronger alternatives: a sparse retriever (BM25~\cite{robertson2009probabilistic}) and advanced dense embedding models (Contriever~\cite{izacard2021unsupervised} and Qwen3-Embedding-8B~\cite{zhang2025qwen3}). 

As shown in Table~\ref{tab:stronger_baselines}, \name consistently outperforms these baselines on both benchmarks. This demonstrates the significant advantage of our history-aware routing mechanism over standard retrieval approaches in complex tool selection.

\begin{table}[ht]
    \centering
    \caption{\textbf{Comparison with stronger baselines.} Performance evaluation of \name against advanced sparse and dense retrieval approaches.}
    \label{tab:stronger_baselines}
    \resizebox{\columnwidth}{!}{%
    \begin{tabular}{lcc}
        \toprule
        \textbf{Method} & \textbf{MCP-Universe (Acc. \%)} & \textbf{MCP-Mark (Acc. \%)} \\
        \midrule
        BM25   & 30.00 & 40.00 \\
        Contriever& 33.66 & 40.00 \\
        Qwen3-Embedding-8B & 44.88 & 42.00 \\
        \midrule
        \textbf{\name (Ours)}                    & \textbf{53.44} & \textbf{60.00} \\
        \bottomrule
    \end{tabular}%
    }
\end{table}

\section{Validity of Simulated Ground Truth}
\label{sec:exp_data_validity}

A critical prerequisite for our framework is ensuring that the simulated training data provides high-quality and reliable supervision. To rigorously quantify the accuracy of the LLM-generated tool calls used as ground truth, we conducted a manual human evaluation. We sampled a random subset of 100 synthesized trajectories and had human experts independently annotate the correct tool selections. The agreement between our synthesized labels and expert annotations was measured using Cohen's Kappa ($\kappa$). As shown in Table~\ref{tab:kappa_score}, the evaluation yields a $\kappa$ score of \textbf{0.93}, indicating near-perfect agreement. This confirms that our self-evolutionary data synthesis mechanism generates highly reliable supervision signals.

\begin{table}[ht]
    \centering
    \caption{\textbf{Reliability of Synthesized Trajectories.} Agreement between LLM-generated tool selections and human expert annotations.}
    \label{tab:kappa_score}
    \begin{tabular}{lc}
        \toprule
        \textbf{Metric} & \textbf{Score} \\
        \midrule
        Cohen's Kappa ($\kappa$) & \textbf{0.93} \\
        \bottomrule
    \end{tabular}
\end{table}

\section{Self-Evolutionary Mutation}
\label{app:mutation}

Table~\ref{tab:mutation-types} presents the taxonomy of mutation types for tools, while Table~\ref{tab:agent-mutation-types} outlines the corresponding mutation strategies designed for agents. Both tables include detailed descriptions and examples

\input{table/mutation_tool}

\section{Agent Route Benchmark}
\label{app:route_bench}
\subsection{Standardized Agent Definition}
\label{sec:agent_def}

In this section, we present the standardized schema used to define agent capabilities within our benchmark. Below are two representative examples: \texttt{SWE\_agent} and \texttt{WebVoyager\_agent}.

\begin{tcolorbox}[
  title={Agent Definition Examples},
  colback=gray!5,
  colframe=black,
  colbacktitle=blue!10!white,
  coltitle=black!85,
  fonttitle=\bfseries,
  breakable,
  enhanced,
  boxrule=0.8pt,
  arc=2mm
]
\small
\begin{lstlisting}[
    basicstyle=\ttfamily\scriptsize, % 使用更小的等宽字体以适应双栏
    breaklines=true,                 % 自动换行
    keywordstyle=\color{blue},       % 关键字颜色（可选）
    stringstyle=\color{teal},        % 字符串颜色（可选）
    columns=fullflexible,
    keepspaces=true
]
[
  {
    "name": "SWE_agent",
    "description": "An autonomous software engineering agent that can understand codebases, modify source code, execute tests, and resolve real GitHub issues through iterative interaction with a development environment.",
    "tools": [
      "find_file", "search_code", "view_file", 
      "edit_file", "apply_patch", "run_tests", 
      "exec_shell", "search_dir", "scroll_up"
    ],
    "inputSchema": {
      "type": "object",
      "properties": {
        "repo_path": {
          "type": "string", 
          "description": "Local filesystem path to the target code repository"
        },
        "issue_description": {
          "type": "string", 
          "description": "Issue title and detailed problem description, typically from a GitHub issue"
        },
        "language": {
          "type": "string", 
          "description": "Primary programming language of the codebase"
        },
        "test_command": {
          "type": "string", 
          "description": "Shell command used to execute the project's test suite"
        },
        "model": {
          "type": "string", 
          "description": "Identifier of the language model used by the agent"
        }
      }
    },
    "tags": ["General"]
  },
  {
    "name": "WebVoyager_agent",
    "description": "A multimodal web agent that uses both text and visual input to interact with real websites end-to-end, completing user instructions by browsing, clicking, typing, and interpreting pages.",
    "tools": [
      "observe_dom", "click", "type_text", 
      "scroll", "navigate"
    ],
    "inputSchema": {
      "type": "object",
      "properties": {
        "instruction": {
          "type": "string", 
          "description": "User's natural language instruction describing the web task to complete"
        },
        "start_url": {
          "type": "string", 
          "description": "Initial webpage URL where the agent begins browsing"
        },
        "multimodal_model": {
          "type": "string", 
          "description": "Identifier of the vision-language model used for perception and reasoning"
        },
        "timeout": {
          "type": "integer", 
          "description": "Maximum allowed wait time for browser interactions"
        }
      }
    },
    "tags": ["General"]
  }
]
\end{lstlisting}
\end{tcolorbox}
\input{table/mutation_agent}
\subsection{Agent Route Benchmark Examples}

\begin{tcolorbox}[
  title={Task Example},
  colback=gray!5,
  colframe=black,
  colbacktitle=blue!10!white,
  coltitle=black!85,
  fonttitle=\bfseries,
  breakable,
  enhanced,
  boxrule=0.8pt,
  arc=2mm
]
\begin{lstlisting}[
  basicstyle=\ttfamily\scriptsize,
  breaklines=true,
  columns=fullflexible,
  keepspaces=true,
  showstringspaces=false,
  upquote=true
]
{
  "system": "You are an Agent Router.
Your task is to analyze the meaning of a user query and select the most relevant agents based on the agents' descriptions and schemas.

Guidelines:
1. Consider both the agent descriptions and input schemas when judging relevance.
2. Use the inputSchema to understand what parameters each agent accepts.
3. Do not infer hidden capabilities or invent agents.
4. Return only one agent that is most relevant.
5. Output strictly in the required format: ["agent_name"], no extra commentary.",
  "user": "Below are examples of the user's past interactions, including queries and system responses:
<history>
User: Initiate a comprehensive code review of the backend systems located at the server path "/var/www/backend" using PHP as the primary language. ...
Assistant: <think>...</think>
<agent_call>code_review_agent{"repo_path": "/var/www/backend", "language": "PHP", "review_criteria": "Performance optimization and best practices adherence", "report_format": "markdown"}</agent_call>
Tool results: ...
Assistant: ## Code Review Report ...

User: I would like to proceed with an analysis of sentiment in the trading data for our company's stocks, symbolized as "COMP", ...
Assistant: <think>...</think>
<agent_call>sentiment_analysis_trading_agent{"market_symbols": ["COMP"], "timeframe": "intraday", "sentiment_threshold": 0.7, "data_source": "auto"}</agent_call>
<agent_call>risk_management_agent{"user_instruction": "Evaluate financial risks ...", "llm_model": "default", "risk_tolerance": "medium"}</agent_call>
Tool results: ...
Assistant: The sentiment analysis for "COMP" indicates ...
</history>

Current user query:
<current query>"Please execute a financial forecast to evaluate the impact on overall business profitability given the recent trading adjustments and risk evaluations. Use a predictive model to analyze the potential outcomes and suggest appropriate policy changes. Ensure that the analysis accounts for the sentiments and risks previously identified, and provide a detailed report on strategic recommendations."</current query>

Available agents:
<agents>[
  {
    "name": "risk_management_agent",
    "description": "The Risk Management Agent specializes in evaluating financial risks based on historical data and current market conditions, providing targeted risk mitigation strategies and recommendations.",
    "tools": ["parse_risk_instruction", "plan_risk_analysis", "generate_risk_model", "execute_risk_assessment", "validate_risk_strategies"],
    "inputSchema": {
      "type": "object",
      "properties": {
        "user_instruction": {"type": "string"},
        "spreadsheet_id": {"type": "string"},
        "llm_model": {"type": "string"},
        "risk_tolerance": {"type": "string"}
      },
      "required": []
    }
  },
  {
    "name": "economy_forecasting_agent",
    "description": "An agent designed to forecast economic trends and guide policy recommendations by simulating various economic scenarios and assessing potential outcomes.",
    "tools": ["simulate_economic_scenario", "forecast_trends", "recommend_policy", "analyze_impact"],
    "inputSchema": {
      "type": "object",
      "properties": {
        "economic_indicators": {"type": "object"},
        "forecast_horizon": {"type": "integer"},
        "policy_options": {"type": "object"},
        "evaluation_criteria": {"type": "object"}
      },
      "required": []
    }
  },
  ... (other agents)
]</agents>

Task:
<task>
Analyze the current query in the context of the user's past queries and agent descriptions.
Return the most relevant agent based on their descriptions and schemas.
</task>

Output requirements:
###
- First, think through your reasoning inside <think></think> tags
- Then output only one agent name as a JSON array
- Format:
<think>
Your reasoning about which agent to select...
</think>

["agent_name"]
###",
  "expected_agent": ["economy_forecasting_agent"]
}
\end{lstlisting}
\end{tcolorbox}

\FloatBarrier
\raggedbottom
\section{Prompt}
\label{app:prompt}

\begin{tcolorbox}[
  title={Self-Evolutionary Mutation Prompt for Tool},
  colback=gray!5,
  colframe=black,
  colbacktitle=blue!10!white,
  coltitle=black!85,
  fonttitle=\bfseries,
  breakable,
  enhanced,
  boxrule=0.8pt,
  arc=2mm
]
\small

\noindent\textbf{\# Role: Expert Tool Designer}

\medskip
You are an expert tool designer specializing in creating innovative software tools through genetic algorithm-inspired mutations. Your expertise includes API design, parameter optimization, and functional enhancement.

\medskip
\noindent\textbf{\#\# Your Task}

\medskip
Perform a \textbf{MUTATION OPERATION} on the given tool to create a new, related but distinct tool that serves a similar domain but with meaningful innovations.

\medskip
\noindent\textbf{\#\# Original Tool Analysis}

\{json.dumps(base\_tool,ensure\_ascii=False,indent=2)\}

\medskip
\noindent\textbf{\#\# Mutation Strategy: \{mutation\_type\}}

\medskip
\noindent\textbf{\#\# Design Requirements}

\medskip
\noindent\textbf{\#\#\# Functional Requirements:}
\begin{itemize}
  \item \textbf{Innovation}: Create meaningful functional differences while maintaining domain relevance
  \item \textbf{Utility}: Ensure the new tool solves a real problem or improves upon existing functionality
  \item \textbf{Compatibility}: Maintain similar complexity level and use case applicability
\end{itemize}

\noindent\textbf{\#\#\# Technical Requirements:}
\begin{itemize}
  \item \textbf{Parameters}: Design intuitive, well-typed parameters following JSON Schema standards
  \item \textbf{Naming}: Use clear, descriptive names that immediately convey purpose
  \item \textbf{Documentation}: Write concise but comprehensive descriptions
  \item \textbf{Validation}: Include appropriate parameter validation and constraints
\end{itemize}

\medskip
\noindent\textbf{\#\#\# Constraints:}
\begin{itemize}
  \item Keep the same domain tags: \{base\_tool.get('tags', [])\}
  \item Avoid direct copying -- ensure meaningful differentiation
  \item Maintain professional tool naming conventions
  \item Focus on practical, implementable functionality
\end{itemize}

\medskip
\noindent\textbf{\#\# Expected Output}

\medskip
Return \textbf{ONLY} valid JSON in this exact format (no markdown, no extra text):

\begin{tcblisting}{
  listing only,
  listing options={
    basicstyle=\ttfamily\footnotesize,
    breaklines=true,
    columns=fullflexible,
    %language=json
  }
}
{
  "name": "descriptive_tool_name",
  "description": "Clear, actionable description of what this tool does and why it's useful",
  "inputSchema": {
    "type": "object",
    "properties": {
      "parameter_name": {
        "type": "appropriate_type",
        "description": "What this parameter does and how to use it",
        "default": "optional_default_value"
      }
    },
    "required": ["list_required_parameters"]
  },
  "tags": {base_tool.get('tags', [])}
}
\end{tcblisting}

\medskip
\noindent\textbf{CRITICAL}: Use only double quotes, no single quotes. No markdown formatting.

\medskip
\noindent\textbf{Note}: Only include a "results" field if the tool produces structured output that requires explicit definition.

\medskip
\noindent\textbf{\#\# Quality Checklist}
\begin{itemize}
  \item[--] Tool name is descriptive and unique
  \item[--] Description clearly explains purpose and value
  \item[--] Parameters are well-designed with proper types
  \item[--] Required parameters are logically necessary
  \item[--] JSON syntax is valid and complete
\end{itemize}

\end{tcolorbox}

\vspace{2em}

\begin{tcolorbox}[
  title={Self-Evolutionary Mutation Prompt for Agent},
  colback=gray!5,
  colframe=black,
  colbacktitle=blue!10!white,
  coltitle=black!85,
  fonttitle=\bfseries,
  breakable,
  enhanced,
  boxrule=0.8pt,
  arc=2mm
]
\small

\noindent\textbf{\# Role: Expert Agent Architect}

\medskip
You are an expert AI agent architect specializing in designing autonomous agents through genetic algorithm-inspired mutations. Your expertise includes agent workflow design, tool orchestration, and capability planning.

\medskip
\noindent\textbf{\#\# Your Task}

\medskip
Perform a \textbf{MUTATION OPERATION} on the given agent to create a new, related but distinct agent that serves a similar purpose but with meaningful innovations in its capabilities and tool composition.

\medskip
\noindent\textbf{\#\# Original Agent Analysis}

\medskip
\textbf{Agent Name}: \{agent\_name\}

\textbf{Description}: \{agent\_description\}

\medskip
\textbf{Tools Used by This Agent}:

\{json.dumps(agent\_tools, ensure\_ascii=False, indent=2)\}

\medskip
\textbf{Agent InputSchema (Parameters)}:

\{json.dumps(agent\_args, ensure\_ascii=False, indent=2)\}

\medskip
\noindent\textbf{\#\# Mutation Strategy: \{mutation\_type\}}

\medskip
\noindent\textbf{\#\# Design Requirements}

\medskip
\noindent\textbf{\#\#\# Agent Design Principles:}
\begin{itemize}
  \item \textbf{Coherent Toolset}: The tools should work together to accomplish the agent's goals
  \item \textbf{Clear Workflow}: The agent should have a logical flow of operations
  \item \textbf{Practical Utility}: The agent should solve real-world problems
  \item \textbf{Tool Synergy}: Tools should complement each other, not duplicate functionality
\end{itemize}

\noindent\textbf{\#\#\# Tool Evolution Guidelines:}
\begin{itemize}
  \item You may ADD new tools that enhance the agent's capabilities
  \item You may MODIFY existing tools to better fit the new agent's purpose
  \item You may REMOVE tools that don't align with the new agent's focus
  \item You may RENAME tools to reflect their new context
  \item Aim for 4--8 tools per agent (not too few, not too many)
\end{itemize}

\noindent\textbf{\#\#\# Naming Convention:}
\begin{itemize}
  \item Agent name MUST end with "\_agent" suffix
  \item Use snake\_case format
  \item Name should clearly indicate the agent's primary function
  \item Example: "code\_review\_agent", "data\_analysis\_agent", "document\_qa\_agent"
\end{itemize}

\noindent\textbf{\#\#\# Tags Guidelines:}
\begin{itemize}
  \item Tags should categorize the agent's primary domain or capability
  \item Use descriptive tags like: "code agent", "search agent", "web agent", "data agent", "research agent", "automation agent", "analysis agent", "multimodal agent", etc.
  \item Can include multiple tags if the agent spans multiple domains
\end{itemize}

\medskip
\noindent\textbf{\#\# Expected Output}

\medskip
Return \textbf{ONLY} valid JSON in this exact format (no markdown, no extra text):

\begin{tcblisting}{
  listing only,
  listing options={
    basicstyle=\ttfamily\scriptsize,
    breaklines=true,
    columns=fullflexible,
    %language=json
  }
}
{
  "name": "descriptive_name_agent",
  "description": "Clear description of what this agent does, its primary use cases, and how it accomplishes its goals",
  "tools": [
    "tool_name_1",
    "tool_name_2",
    "tool_name_3"
  ],
  "inputSchema": {
    "type": "object",
    "properties": {
      "parameter_name": {
        "type": "appropriate_type",
        "description": "Detailed description of what this parameter configures for the agent"
      }
    }
  },
  "tags": ["category agent"]
}
\end{tcblisting}

\medskip
\noindent\textbf{CRITICAL REQUIREMENTS:}
\begin{itemize}
  \item Agent name MUST end with "\_agent"
  \item Use only double quotes, no single quotes
  \item No markdown formatting
  \item Tools array should contain 4--8 tool names
  \item Each tool name should be descriptive and use snake\_case
  \item Tags should be descriptive category labels (e.g., "code agent", "search agent", "web agent")
  \item Each parameter in inputSchema.properties MUST have a detailed "description" field
\end{itemize}

\medskip
\noindent\textbf{\#\# Quality Checklist}
\begin{itemize}
  \item[--] Agent name ends with "\_agent" and clearly describes purpose
  \item[--] Description explains the agent's workflow and capabilities
  \item[--] Tools form a coherent set that enables the agent's goals
  \item[--] Tools are appropriately evolved from the original (not just copied)
  \item[--] Parameters make sense for configuring this agent
  \item[--] Each parameter has a clear, detailed description in inputSchema
  \item[--] Tags accurately categorize the agent's domain
  \item[--] JSON syntax is valid and complete
\end{itemize}

\end{tcolorbox}

\section{Confidence Intervals and Variance Analysis}
\label{sec:appendix_variance}

To rigorously demonstrate the reliability of our results and address the robustness of our evaluations, we calculate and report the Standard Deviation (STD) across 5 independent runs for our main experiments. 

Table~\ref{tab:mcp_overall} updates the overall benchmark performance, presenting the variance for the final accuracy on MCP-Universe and MCP-Mark. Furthermore, Table~\ref{tab:robustness_noise} provides the comprehensive robustness and scalability analysis with full confidence intervals under all experimental settings. These results confirm that \name maintains statistically significant improvements over the baselines.

\begin{table*}[ht]
\centering
\caption{\textbf{Overall Performance.} Accuracy (\%) with standard deviation across 5 independent runs on the MCP-Universe and MCP-Mark benchmarks. The best result is marked in \textbf{bold} and the second best is \underline{underlined}.}
\label{tab:mcp_overall}
\begin{tabular}{lcc}
\toprule
\textbf{Method} & \textbf{MCP-Universe (Overall)} & \textbf{MCP-Mark (Overall)} \\
\midrule

\rowcolor{gray!12}
\multicolumn{3}{c}{\textit{Embedding-based Retrieval}}\\
Text-Emb-3-Large (Q)   & 40.95 $\pm$ 1.43 & 26.00 $\pm$ 2.45 \\
Text-Emb-3-Large (Q+H) & 35.20 $\pm$ 1.51 & 20.00 $\pm$ 2.00 \\
all-MiniLM-L6-v2 (Q)   & 40.67 $\pm$ 1.95 & 26.00 $\pm$ 3.46 \\
all-MiniLM-L6-v2 (Q+H) & 43.62 $\pm$ 1.25 & 32.00 $\pm$ 3.46 \\
\midrule

\rowcolor{gray!12}
\multicolumn{3}{c}{\textit{ReAct Agents}}\\
ReAct (Gemini-2.5-Pro) & 41.80 $\pm$ 1.65 & 40.00 $\pm$ 2.00 \\
\midrule

\rowcolor{gray!12}
\multicolumn{3}{c}{\textit{LLM-based Routers}}\\
Qwen3-8B               & 46.14 $\pm$ 1.56 & 42.00 $\pm$ 2.00 \\
GLM-4.5                & 46.42 $\pm$ 0.99 & 38.00 $\pm$ 3.46 \\
DeepSeek-V3.2          & 44.74 $\pm$ 1.48 & 42.00 $\pm$ 2.00 \\
Claude-Sonnet-4        & 48.25 $\pm$ 1.93 & 44.00 $\pm$ 2.45 \\
Gemini-2.5-Pro         & \underline{49.79 $\pm$ 1.36} & 46.00 $\pm$ 2.45 \\
Gemini-2.5-flash       & 46.98 $\pm$ 1.66 & \underline{48.00 $\pm$ 2.83} \\
GPT-4.1                & 44.60 $\pm$ 1.45 & 36.00 $\pm$ 2.00 \\
GPT-4o                 & 47.41 $\pm$ 1.50 & 38.00 $\pm$ 2.83 \\
\rowcolor{blue!20}
\textbf{\name (Ours)}  & \textbf{53.44 $\pm$ 0.57} & \textbf{60.00 $\pm$ 2.00} \\
\bottomrule
\end{tabular}
\end{table*}

\begin{table*}[ht]
\centering
\small
\setlength{\tabcolsep}{3.5pt}
\caption{\textbf{Robustness and Scalability Analysis.} We evaluate model performance on MCP-Universe and MCP-Mark under four settings: \textbf{Clean} (single-server baseline), \textbf{Multi} (merged multi-server tools setting), \textbf{+Mutation} (adding Self-Evolutionary mutation tools), and \textbf{+LiveMCP} (adding real external tools). The best result is marked in \textbf{bold} and the second best is \underline{underlined}.}
\label{tab:robustness_noise}
\resizebox{\textwidth}{!}{
\begin{tabular}{l cccc cccc}
\toprule
\multirow{2}{*}{\textbf{Method}} &
\multicolumn{4}{c}{\textbf{MCP-Universe}} &
\multicolumn{4}{c}{\textbf{MCP-Mark}} \\
\cmidrule(lr){2-5} \cmidrule(lr){6-9}
 & Clean & Multi & +Mutation & +LiveMCP & Clean & Multi & +Mutation & +LiveMCP \\
\midrule

\rowcolor{gray!12}
\multicolumn{9}{c}{\textit{Embedding-based Retrieval}}\\
Text-Emb-3-Large (Q)       & 40.95 $\pm$ 1.43 & 40.11 $\pm$ 1.89 & 39.69 $\pm$ 1.38 & 38.99 $\pm$ 1.27 & 26.00 $\pm$ 2.45 & 18.00 $\pm$ 3.16 & 20.00 $\pm$ 2.83 & 14.00 $\pm$ 2.83 \\
Text-Emb-3-Large (Q+H)     & 35.20 $\pm$ 1.51 & 34.22 $\pm$ 1.96 & 34.36 $\pm$ 1.38 & 33.24 $\pm$ 1.95 & 20.00 $\pm$ 2.00 & 14.00 $\pm$ 3.46 & 12.00 $\pm$ 2.00 & 10.00 $\pm$ 3.46 \\
all-MiniLM-L6-v2 (Q)       & 40.67 $\pm$ 1.95 & 39.69 $\pm$ 0.59 & 39.83 $\pm$ 1.99 & 38.57 $\pm$ 1.97 & 26.00 $\pm$ 3.46 & 20.00 $\pm$ 2.45 & 22.00 $\pm$ 2.83 & 14.00 $\pm$ 2.45 \\
all-MiniLM-L6-v2 (Q+H)     & 43.62 $\pm$ 1.25 & 42.92 $\pm$ 1.17 & 42.78 $\pm$ 1.69 & 41.80 $\pm$ 0.98 & 32.00 $\pm$ 3.46 & 26.00 $\pm$ 2.45 & 20.00 $\pm$ 2.83 & 18.00 $\pm$ 3.46 \\
\midrule

\rowcolor{gray!12}
\multicolumn{9}{c}{\textit{ReAct Agents}}\\
ReAct (Gemini-2.5-Pro)     & 41.80 $\pm$ 1.65 & 36.47 $\pm$ 1.02 & 40.95 $\pm$ 1.39 & 40.11 $\pm$ 1.36 & 40.00 $\pm$ 2.00 & 32.00 $\pm$ 2.83 & 26.00 $\pm$ 3.16 & 20.00 $\pm$ 1.41 \\
\midrule

\rowcolor{gray!12}
\multicolumn{9}{c}{\textit{LLM-based Routers}}\\
Qwen3-8B                   & 46.14 $\pm$ 1.56 & 45.30 $\pm$ 1.46 & 44.88 $\pm$ 1.83 & 44.46 $\pm$ 1.75 & 42.00 $\pm$ 2.00 & 38.00 $\pm$ 2.45 & 32.00 $\pm$ 1.41 & 30.00 $\pm$ 2.83 \\
GLM-4.5                    & 46.42 $\pm$ 0.99 & 45.44 $\pm$ 1.70 & 45.30 $\pm$ 0.75 & 44.32 $\pm$ 1.61 & 38.00 $\pm$ 3.46 & 32.00 $\pm$ 3.16 & 30.00 $\pm$ 3.16 & 26.00 $\pm$ 3.46 \\
DeepSeek-V3.2              & 44.74 $\pm$ 1.48 & 43.76 $\pm$ 1.94 & 44.04 $\pm$ 1.30 & 43.06 $\pm$ 0.95 & 42.00 $\pm$ 2.00 & 38.00 $\pm$ 3.16 & 36.00 $\pm$ 2.83 & 30.00 $\pm$ 2.83 \\
Claude-Sonnet-4          & 48.25 $\pm$ 1.93 & 47.41 $\pm$ 1.59 & 47.12 $\pm$ 1.49 & 46.84 $\pm$ 1.97 & 44.00 $\pm$ 2.45 & \underline{40.00 $\pm$ 3.16} & 32.00 $\pm$ 3.46 & 30.00 $\pm$ 2.83 \\
Gemini-2.5-Pro             & \underline{49.79 $\pm$ 1.36} & \underline{49.09 $\pm$ 1.81} & \underline{48.81 $\pm$ 1.90} & \underline{48.11 $\pm$ 1.34} & 46.00 $\pm$ 2.45 & \underline{40.00 $\pm$ 2.45} & 36.00 $\pm$ 3.16 & 32.00 $\pm$ 2.83 \\
Gemini-2.5-flash           & 46.98 $\pm$ 1.66 & 46.28 $\pm$ 1.89 & 46.00 $\pm$ 0.82 & 43.76 $\pm$ 1.76 & \underline{48.00 $\pm$ 2.83} & \underline{40.00 $\pm$ 2.00} & \underline{44.00 $\pm$ 2.45} & \underline{36.00 $\pm$ 3.46} \\
GPT-4.1                    & 44.60 $\pm$ 1.45 & 43.90 $\pm$ 1.02 & 43.48 $\pm$ 1.44 & 42.64 $\pm$ 1.48 & 36.00 $\pm$ 2.00 & 32.00 $\pm$ 2.00 & 30.00 $\pm$ 2.00 & 28.00 $\pm$ 2.83 \\
GPT-4o                     & 47.41 $\pm$ 1.50 & 46.56 $\pm$ 1.62 & 46.42 $\pm$ 0.74 & 45.86 $\pm$ 1.75 & 38.00 $\pm$ 2.83 & 34.00 $\pm$ 3.16 & 32.00 $\pm$ 2.83 & 28.00 $\pm$ 2.45 \\
\rowcolor{blue!20}
\textbf{\name (Ours)}      & \textbf{53.44 $\pm$ 0.57} & \textbf{53.02 $\pm$ 1.21} & \textbf{53.02 $\pm$ 0.87} & \textbf{52.59 $\pm$ 0.50} & \textbf{60.00 $\pm$ 2.00} & \textbf{56.00 $\pm$ 1.41} & \textbf{54.00 $\pm$ 2.45} & \textbf{56.00 $\pm$ 2.83} \\
\bottomrule
\end{tabular}
}
\end{table*}


%% file: table/mutation_tool.tex
\begin{table}[!htbp]
\centering
\small
\caption{Taxonomy of Tool Mutation Types}
\label{tab:mutation-types}
\setlength{\tabcolsep}{6pt}
\renewcommand{\arraystretch}{1.05}
\begin{tabular}{p{0.32\columnwidth} p{0.64\columnwidth}}
\toprule
\textbf{Mutation Type} & \textbf{Description \& Examples} \\
\midrule

\textbf{Usage Extension}
& Apply the tool's core logic to related new scenarios or domains. \textit{Example}:
\texttt{analyze\_code\_quality} $\to$ \texttt{analyze\_document\_quality} (applies code analysis concepts to documents). \\
\midrule

\textbf{Function Enhancement}
& Substantially expand the tool's capabilities to enable entirely new use cases while maintaining the core purpose (add 2+ major user-visible features). \textit{Example}:
\texttt{compress\_image} $\to$ \texttt{image\_optimization\_suite} (adds format conversion + batch processing + quality presets + metadata editing). \\
\midrule

\textbf{Workflow Chain}
& Create a tool that works immediately before or after the original tool in a workflow, providing better inputs or processing outputs. \textit{Example}:
\texttt{search\_web} $\to$ \texttt{prepare\_search\_keywords} (pre-processes queries) or \texttt{summarize\_search\_results} (post-processes results). \\
\midrule

\textbf{Helper Tool}
& Create an independent supporting tool that enhances the ecosystem around the original tool. \textit{Example}:
\texttt{create\_chart} $\to$ \texttt{validate\_chart\_data} (checks data format before charting) or \texttt{suggest\_chart\_colors} (recommends color schemes). \\
\midrule

\textbf{Parameter Redesign}
& Modify the tool's parameter structure to enable different input patterns or interaction approaches. Focus on meaningful parameter changes that shift how users provide data or configure behavior. \textit{Example}:
\texttt{get\_user(user\_id: string)} $\to$ \texttt{query\_users(filters: object, sort: string, limit: number)} (from single lookup to flexible querying). \\
\bottomrule
\end{tabular}
\end{table}

%% file: table/mutation_agent.tex
\begin{table}[!htbp]
\centering
\small
\caption{Taxonomy of Agent Mutation Types}
\label{tab:agent-mutation-types}
\setlength{\tabcolsep}{6pt}
\renewcommand{\arraystretch}{1.2} 
\begin{tabular}{p{0.32\columnwidth} p{0.64\columnwidth}}
\toprule
\textbf{Mutation Type} & \textbf{Description \& Examples} \\
\midrule

\textbf{Domain Transfer}
& Apply the agent's architecture and workflow to a different but related domain. \textit{Example}:
\texttt{SWE\_agent} $\to$ \texttt{doc\_review\_agent} (adapts the edit/search/validate pattern from code to documents). \\
\midrule

\textbf{Capability Enhancement}
& Substantially expand the agent's capabilities by adding new tools and extending its scope. \textit{Example}:
\texttt{code\_search\_agent} $\to$ \texttt{code\_intelligence\_agent} (adds semantic analysis, dependency tracking, and refactoring suggestions). \\
\midrule

\textbf{Workflow Specialization}
& Create a more focused agent that specializes in a subset of the original agent's workflow. \textit{Example}:
\texttt{full\_stack\_dev\_agent} $\to$ \texttt{api\_testing\_agent} (focuses exclusively on API testing with specialized validation tools). \\
\midrule

\textbf{Tool Composition}
& Recombine and restructure the agent's tools to create new workflow patterns. \textit{Example}:
\texttt{data\_pipeline\_agent} $\to$ \texttt{realtime\_streaming\_agent} (reorganizes batch processing tools into streaming-compatible tools). \\
\midrule

\textbf{Scenario Adaptation}
& Adapt the agent to handle different use case scenarios or user contexts. \textit{Example}:
\texttt{general\_qa\_agent} $\to$ \texttt{customer\_support\_agent} (adapts general QA capabilities specifically for customer service scenarios). \\
\bottomrule
\end{tabular}
\end{table}

%% file: main.bib
@article{guo2025deepseek,
  title={Deepseek-r1: Incentivizing reasoning capability in llms via reinforcement learning},
  author={Guo, Daya and Yang, Dejian and Zhang, Haowei and Song, Junxiao and Zhang, Ruoyu and Xu, Runxin and Zhu, Qihao and Ma, Shirong and Wang, Peiyi and Bi, Xiao and others},
  journal={arXiv preprint arXiv:2501.12948},
  year={2025}
}

@article{achiam2023gpt,
  title={Gpt-4 technical report},
  author={Achiam, Josh and Adler, Steven and Agarwal, Sandhini and Ahmad, Lama and Akkaya, Ilge and Aleman, Florencia Leoni and Almeida, Diogo and Altenschmidt, Janko and Altman, Sam and Anadkat, Shyamal and others},
  journal={arXiv preprint arXiv:2303.08774},
  year={2023}
}

@article{comanici2025gemini,
  title={Gemini 2.5: Pushing the frontier with advanced reasoning, multimodality, long context, and next generation agentic capabilities},
  author={Comanici, Gheorghe and Bieber, Eric and Schaekermann, Mike and Pasupat, Ice and Sachdeva, Noveen and Dhillon, Inderjit and Blistein, Marcel and Ram, Ori and Zhang, Dan and Rosen, Evan and others},
  journal={arXiv preprint arXiv:2507.06261},
  year={2025}
}

@misc{anthropic2024introducing,
  author       = {Anthropic},
  title        = {{Introducing the Model Context Protocol}},
  year         = {2024},
  month        = nov,
  howpublished = {\url{https://www.anthropic.com/news/model-context-protocol}},
}

@misc{anthropic2025claude4,
  author       = {Anthropic},
  title        = {{Introducing Claude 4}},
  year         = {2025},
  month        = may,
  howpublished = {\url{https://www.anthropic.com/news/claude-4}},
}

@article{yang2025agentic,
  title={Agentic web: Weaving the next web with ai agents},
  author={Yang, Yingxuan and Ma, Mulei and Huang, Yuxuan and Chai, Huacan and Gong, Chenyu and Geng, Haoran and Zhou, Yuanjian and Wen, Ying and Fang, Meng and Chen, Muhao and others},
  journal={arXiv preprint arXiv:2507.21206},
  year={2025}
}

@inproceedings{yao2023react,
  title={ReAct: Synergizing Reasoning and Acting in Language Models},
  author={Yao, Shunyu and Zhao, Dian and Yu, Jeffrey and Shafran, Nan and Narasimhan, Karthik and Cao, Yuan},
  booktitle={International Conference on Learning Representations},
  year={2023},
  url={https://openreview.net/forum?id=WE_vluYUL-X}
}

@article{schick2023toolformer,
  title={Toolformer: Language Models Can Teach Themselves to Use Tools},
  author={Schick, Timo and Dwivedi-Yu, Jane and Dess{\`{i}}, Roberto and Raileanu, Roberta and Lomeli, Maria and Zettlemoyer, Luke and Cancedda, Nicola and Scialom, Thomas},
  journal={arXiv preprint arXiv:2302.04761},
  year={2023},
  url={https://arxiv.org/abs/2302.04761}
}

@article{shen2023hugginggpt,
  title={HuggingGPT: Solving AI Tasks with ChatGPT and its Friends in Hugging Face},
  author={Shen, Yongliang and Song, Kaitao and Tan, Xu and Li, Dongsheng and Lu, Weiming and Zhuang, Yueting},
  journal={arXiv preprint arXiv:2303.17580},
  year={2023},
  url={https://arxiv.org/abs/2303.17580}
}

@article{patil2023gorilla,
  title={Gorilla: Large Language Model Connected with Massive APIs},
  author={Patil, Shishir G and Zhang, Tianjun and Wang, Xin and Gonzalez, Joseph E},
  journal={arXiv preprint arXiv:2305.15334},
  year={2023},
  url={https://arxiv.org/abs/2305.15334}
}

@article{qin2023toolllm,
  title={ToolLLM: Facilitating Large Language Models to Master 16000+ Real-world APIs},
  author={Qin, Yujia and Liang, Shihao and Ye, Yining and Zhu, Kunlun and Yan, Lan and Lu, Yaxi and Lin, Yankai and Cong, Xin and Tang, Xiangru and Qian, Bill and others},
  journal={arXiv preprint arXiv:2307.16789},
  year={2023},
  url={https://arxiv.org/abs/2307.16789}
}

@article{zhang2024toolnet,
  title={ToolNet: Connecting Large Language Models with Massive Tools via Tool Graph},
  author={Zhang, Xukun and Zhu, Zhiyuan and Wang, Mingyu and Wang, Lingfei and Li, Haoran and Zhang, Jingjing and Li, Dongsheng},
  journal={arXiv preprint arXiv:2403.00839},
  year={2024},
  url={https://arxiv.org/abs/2403.00839}
}

@article{song2023restgpt,
  title={RestGPT: Connecting Large Language Models with Real-World RESTful APIs},
  author={Song, Yifan and Xiong, Weimin and Zhu, Dawei and Li, Cheng and Wang, Ke and Tian, Ye and Li, Sujian},
  journal={arXiv preprint arXiv:2306.06624},
  year={2023},
  url={https://arxiv.org/abs/2306.06624}
}

@inproceedings{li2023apibank,
  title={API-Bank: A Comprehensive Benchmark for Tool-Augmented LLMs},
  author={Li, Minghao and Song, Feifan and Yu, Bowen and Yu, Haiyang and Li, Zhoujun and Huang, Fei and Li, Yongbin},
  booktitle={Proceedings of the 2023 Conference on Empirical Methods in Natural Language Processing},
  pages={3102--3116},
  year={2023},
  publisher={Association for Computational Linguistics},
  url={https://aclanthology.org/2023.emnlp-main.187}
}

@article{yang2024autotool,
  title={AutoTool: Efficient Tool Selection for Large Language Model Agents},
  author={Yang, Jian and Wang, Zhao and Li, Yuxiang and Chen, Hao and Wang, Ke and Li, Yingwei and He, Jingrui},
  journal={arXiv preprint arXiv:2511.14650},
  year={2024},
  url={https://arxiv.org/abs/2511.14650}
}

@article{scalemcp,
  title={ScaleMCP: Dynamic and Auto-Synchronizing Model Context Protocol Tools for LLM Agents},
  author={Lumer, Elias and Gulati, Anmol and Subbiah, Vamse Kumar and Basavaraju, Pradeep Honaganahalli and Burke, James A},
  journal={arXiv preprint arXiv:2505.06416},
  year={2025}
}

@article{li2024toolsandbox,
  title={ToolSandbox: A Stateful, Conversational, Interactive Evaluation Benchmark for LLM Tool Use Capabilities},
  author={Lu, Jiarui and Holleis, Thomas and Zhang, Yizhe and Aumayer, Bernhard and Nan, Feng and Bai, Felix and Ma, Shuang and Ma, Shen and Li, Mengyu and Yin, Guoli and Wang, Zirui and Pang, Ruoming},
  journal={arXiv preprint arXiv:2408.04682},
  year={2024},
  url={https://arxiv.org/abs/2408.04682}
}

@inproceedings{yan2024bfcl,
  title={The Berkeley Function Calling Leaderboard (BFCL): From Tool Use to Agentic Evaluation of Large Language Models},
  author={Patil, Shishir G. and Mao, Huanzhi and Ji, Charlie Cheng-Jie and Yan, Fanjia and Suresh, Vishnu and Stoica, Ion and Gonzalez, Joseph E.},
  booktitle={Proceedings of the 42nd International Conference on Machine Learning},
  year={2025}
}

@article{chen2024teval,
  title={T-Eval: Evaluating the Tool Utilization Capability of Large Language Models Step by Step},
  author={Chen, Zehui and Du, Weihua and Zhang, Wenwei and Liu, Kuikun and Liu, Jiangning and Zheng, Miao and Zhuo, Jingming and Zhang, Songyang and Lin, Dahua and Chen, Kai and Zhao, Feng},
  journal={arXiv preprint arXiv:2312.14033},
  year={2024},
  url={https://arxiv.org/abs/2312.14033}
}

@article{chen2024dtdr,
  title={Dynamic Tool Dependency Retrieval for Efficient Function Calling},
  author={Patel, Bhrij and Belli, Davide and Jalalirad, Amir and Arnold, Maximilian and Ermovol, Aleksandr and Major, Bence},
  journal={arXiv preprint arXiv:2512.17052},
  year={2025},
  url={https://arxiv.org/abs/2512.17052}
}

@article{cai2024toolchain,
  title={ToolChain*: Efficient Action Space Navigation in Large Language Models with A* Search},
  author={Zhuang, Yuchen and Chen, Xiang and Yu, Tong and Mitra, Saayan and Bursztyn, Victor and Rossi, Ryan A. and Sarkhel, Somdeb and Zhang, Chao},
  journal={arXiv preprint arXiv:2310.13227},
  year={2023},
  url={https://arxiv.org/abs/2310.13227}
}

@misc{hu2024sitgraph,
  title={SIT-Graph: State Integrated Tool Graph for Multi-Turn Agents}, 
  author={Li, Sijia and Huang, Yuchen and Liu, Zifan and Li, Zijian and Fu, Jingjing and Song, Lei and Bian, Jiang and Zhang, Jun and Wang, Rui},
  year={2025},
  eprint={2512.07287},
  archivePrefix={arXiv},
  primaryClass={cs.LG},
  url={https://arxiv.org/abs/2512.07287}
}

@article{yang2024swe,
  title={Swe-agent: Agent-computer interfaces enable automated software engineering},
  author={Yang, John and Jimenez, Carlos E and Wettig, Alexander and Lieret, Kilian and Yao, Shunyu and Narasimhan, Karthik and Press, Ofir},
  journal={Advances in Neural Information Processing Systems},
  volume={37},
  pages={50528--50652},
  year={2024}
}

@misc{guo2025octopus,
      title={Octopus: Agentic Multimodal Reasoning with Six-Capability Orchestration}, 
      author={Yifu Guo and Zishan Xu and Zhiyuan Yao and Yuquan Lu and Jiaye Lin and Sen Hu and Zhenheng Tang and Huacan Wang and Ronghao Chen},
      year={2025},
      eprint={2511.15351},
      archivePrefix={arXiv},
      primaryClass={cs.AI},
      url={https://arxiv.org/abs/2511.15351}, 
}

@article{li2025search,
  title={Search-o1: Agentic search-enhanced large reasoning models},
  author={Li, Xiaoxi and Dong, Guanting and Jin, Jiajie and Zhang, Yuyao and Zhou, Yujia and Zhu, Yutao and Zhang, Peitian and Dou, Zhicheng},
  journal={arXiv preprint arXiv:2501.05366},
  year={2025}
}

@article{fang2025comprehensive,
  title={A comprehensive survey of self-evolving ai agents: A new paradigm bridging foundation models and lifelong agentic systems},
  author={Fang, Jinyuan and Peng, Yanwen and Zhang, Xi and Wang, Yingxu and Yi, Xinhao and Zhang, Guibin and Xu, Yi and Wu, Bin and Liu, Siwei and Li, Zihao and others},
  journal={arXiv preprint arXiv:2508.07407},
  year={2025}
}

@article{tran2025multi,
  title={Multi-agent collaboration mechanisms: A survey of llms},
  author={Tran, Khanh-Tung and Dao, Dung and Nguyen, Minh-Duong and Pham, Quoc-Viet and O'Sullivan, Barry and Nguyen, Hoang D},
  journal={arXiv preprint arXiv:2501.06322},
  year={2025}
}

@article{sapkota2025ai,
  title={Ai agents vs. agentic ai: A conceptual taxonomy, applications and challenges},
  author={Sapkota, Ranjan and Roumeliotis, Konstantinos I and Karkee, Manoj},
  journal={arXiv preprint arXiv:2505.10468},
  year={2025}
}

@article{gan2025rag,
  title={Rag-mcp: Mitigating prompt bloat in llm tool selection via retrieval-augmented generation},
  author={Gan, Tiantian and Sun, Qiyao},
  journal={arXiv preprint arXiv:2505.03275},
  year={2025}
}

@article{liu2024toolace,
  title={Toolace: Winning the points of llm function calling},
  author={Liu, Weiwen and Huang, Xu and Zeng, Xingshan and Hao, Xinlong and Yu, Shuai and Li, Dexun and Wang, Shuai and Gan, Weinan and Liu, Zhengying and Yu, Yuanqing and others},
  journal={arXiv preprint arXiv:2409.00920},
  year={2024}
}

@inproceedings{wang2025toolflow,
  title={Toolflow: Boosting llm tool-calling through natural and coherent dialogue synthesis},
  author={Wang, Zezhong and Zeng, Xingshan and Liu, Weiwen and Li, Liangyou and Wang, Yasheng and Shang, Lifeng and Jiang, Xin and Liu, Qun and Wong, Kam-Fai},
  booktitle={Proceedings of the 2025 Conference of the Nations of the Americas Chapter of the Association for Computational Linguistics: Human Language Technologies (Volume 1: Long Papers)},
  pages={4246--4263},
  year={2025}
}

@article{luo2025mcp,
  title={Mcp-universe: Benchmarking large language models with real-world model context protocol servers},
  author={Luo, Ziyang and Shen, Zhiqi and Yang, Wenzhuo and Zhao, Zirui and Jwalapuram, Prathyusha and Saha, Amrita and Sahoo, Doyen and Savarese, Silvio and Xiong, Caiming and Li, Junnan},
  journal={arXiv preprint arXiv:2508.14704},
  year={2025}
}

@article{wu2025mcpmark,
  title={MCPMark: A Benchmark for Stress-Testing Realistic and Comprehensive MCP Use},
  author={Wu, Zijian and Liu, Xiangyan and Zhang, Xinyuan and Chen, Lingjun and Meng, Fanqing and Du, Lingxiao and Zhao, Yiran and Zhang, Fanshi and Ye, Yaoqi and Wang, Jiawei and others},
  journal={arXiv preprint arXiv:2509.24002},
  year={2025}
}

@article{mo2025livemcpbench,
  title={Livemcpbench: Can agents navigate an ocean of mcp tools?},
  author={Mo, Guozhao and Zhong, Wenliang and Chen, Jiawei and Chen, Xuanang and Lu, Yaojie and Lin, Hongyu and He, Ben and Han, Xianpei and Sun, Le},
  journal={arXiv preprint arXiv:2508.01780},
  year={2025}
}

@misc{shi2025aimefullyautonomousmultiagentframework,
      title={Aime: Towards Fully-Autonomous Multi-Agent Framework}, 
      author={Yexuan Shi and Mingyu Wang and Yunxiang Cao and Hongjie Lai and Junjian Lan and Xin Han and Yu Wang and Jie Geng and Zhenan Li and Zihao Xia and Xiang Chen and Chen Li and Jian Xu and Wenbo Duan and Yuanshuo Zhu},
      year={2025},
      eprint={2507.11988},
      archivePrefix={arXiv},
      primaryClass={cs.AI},
      url={https://arxiv.org/abs/2507.11988}, 
}

@misc{hong2025deepeyesv2,
      title={DeepEyesV2: Toward Agentic Multimodal Model}, 
      author={Jack Hong and Chenxiao Zhao and ChengLin Zhu and Weiheng Lu and Guohai Xu and Xing Yu},
      year={2025},
      eprint={2511.05271},
      archivePrefix={arXiv},
      primaryClass={cs.CV},
      url={https://arxiv.org/abs/2511.05271}, 
}

@misc{lù2025buildwebagentsagents,
      title={Build the web for agents, not agents for the web}, 
      author={Xing Han Lù and Gaurav Kamath and Marius Mosbach and Siva Reddy},
      year={2025},
      eprint={2506.10953},
      archivePrefix={arXiv},
      primaryClass={cs.LG},
      url={https://arxiv.org/abs/2506.10953}, 
}

@misc{petrova2025semanticwebmasagentic,
      title={From Semantic Web and MAS to Agentic AI: A Unified Narrative of the Web of Agents}, 
      author={Tatiana Petrova and Boris Bliznioukov and Aleksandr Puzikov and Radu State},
      year={2025},
      eprint={2507.10644},
      archivePrefix={arXiv},
      primaryClass={cs.AI},
      url={https://arxiv.org/abs/2507.10644}, 
}

@article{fan2025mcptoolbench,
  title={MCPToolBench++: A Large Scale AI Agent Model Context Protocol MCP Tool Use Benchmark},
  author={Fan, Shiqing and Ding, Xichen and Zhang, Liang and Mo, Linjian},
  journal={arXiv preprint arXiv:2508.07575},
  year={2025}
}

@misc{guo2025mcpagentbench,
  title={MCP-AgentBench: Evaluating Real-World Language Agent Performance with MCP-Mediated Tools},
  author={Zikang Guo and Benfeng Xu and Chiwei Zhu and Wentao Hong and Xiaorui Wang and Zhendong Mao},
  year={2025},
  eprint={2509.09734},
  archivePrefix={arXiv},
  primaryClass={cs.AI},
  url={https://arxiv.org/abs/2509.09734}
}

@misc{gao2025mcpradar,
  title={MCP-RADAR: A Multi-Dimensional Benchmark for Evaluating Tool Use Capabilities in Large Language Models},
  author={Xuanqi Gao and Siyi Xie and Juan Zhai and Shiqing Ma and Chao Shen},
  year={2025},
  eprint={2505.16700},
  archivePrefix={arXiv},
  primaryClass={cs.AI},
  url={https://arxiv.org/abs/2505.16700}
}

@article{yang2025qwen3,
  title={Qwen3 technical report},
  author={Yang, An and Li, Anfeng and Yang, Baosong and Zhang, Beichen and Hui, Binyuan and Zheng, Bo and Yu, Bowen and Gao, Chang and Huang, Chengen and Lv, Chenxu and others},
  journal={arXiv preprint arXiv:2505.09388},
  year={2025}
}

@misc{seagent,
      title={SE-Agent: Self-Evolution Trajectory Optimization in Multi-Step Reasoning with LLM-Based Agents}, 
      author={Jiaye Lin and Yifu Guo and Yuzhen Han and Sen Hu and Ziyi Ni and Licheng Wang and Mingguang Chen and Hongzhang Liu and Ronghao Chen and Yangfan He and Daxin Jiang and Binxing Jiao and Chen Hu and Huacan Wang},
      year={2025},
      eprint={2508.02085},
      archivePrefix={arXiv},
      primaryClass={cs.AI},
      url={https://arxiv.org/abs/2508.02085}, 
}

@article{hu2022lora,
  title={Lora: Low-rank adaptation of large language models.},
  author={Hu, Edward J and Shen, Yelong and Wallis, Phillip and Allen-Zhu, Zeyuan and Li, Yuanzhi and Wang, Shean and Wang, Lu and Chen, Weizhu and others},
  journal={ICLR},
  volume={1},
  number={2},
  pages={3},
  year={2022}
}

@article{hurst2024gpt,
  title={Gpt-4o system card},
  author={Hurst, Aaron and Lerer, Adam and Goucher, Adam P and Perelman, Adam and Ramesh, Aditya and Clark, Aidan and Ostrow, AJ and Welihinda, Akila and Hayes, Alan and Radford, Alec and others},
  journal={arXiv preprint arXiv:2410.21276},
  year={2024}
}

@article{novikov2025alphaevolve,
  title={Alphaevolve: A coding agent for scientific and algorithmic discovery},
  author={Novikov, Alexander and V{\~u}, Ng{\^a}n and Eisenberger, Marvin and Dupont, Emilien and Huang, Po-Sen and Wagner, Adam Zsolt and Shirobokov, Sergey and Kozlovskii, Borislav and Ruiz, Francisco JR and Mehrabian, Abbas and others},
  journal={arXiv preprint arXiv:2506.13131},
  year={2025}
}

@book{robertson2009probabilistic,
  title={The probabilistic relevance framework: BM25 and beyond},
  author={Robertson, Stephen and Zaragoza, Hugo},
  volume={4},
  year={2009},
  publisher={Now Publishers Inc}
}

@article{izacard2021unsupervised,
  title={Unsupervised dense information retrieval with contrastive learning},
  author={Izacard, Gautier and Caron, Mathilde and Hosseini, Lucas and Riedel, Sebastian and Bojanowski, Piotr and Joulin, Armand and Grave, Edouard},
  journal={arXiv preprint arXiv:2112.09118},
  year={2021}
}

@article{zhang2025qwen3,
  title={Qwen3 embedding: Advancing text embedding and reranking through foundation models},
  author={Zhang, Yanzhao and Li, Mingxin and Long, Dingkun and Zhang, Xin and Lin, Huan and Yang, Baosong and Xie, Pengjun and Yang, An and Liu, Dayiheng and Lin, Junyang and others},
  journal={arXiv preprint arXiv:2506.05176},
  year={2025}
}

@misc{xu2026unlockingimplicitexperiencesynthesizing,
      title={Unlocking Implicit Experience: Synthesizing Tool-Use Trajectories from Text}, 
      author={Zhihao Xu and Rumei Li and Jiahuan Li and Rongxiang Weng and Jingang Wang and Xunliang Cai and Xiting Wang},
      year={2026},
      eprint={2601.10355},
      archivePrefix={arXiv},
      primaryClass={cs.CL},
      url={https://arxiv.org/abs/2601.10355}, 
}

@article{chang2025grail,
  title={GRAIL: Learning to Interact with Large Knowledge Graphs for Retrieval Augmented Reasoning},
  author={Chang, Ge and Su, Jinbo and Liu, Jiacheng and Yang, Pengfei and Shang, Yuhao and Zheng, Huiwen and Ma, Hongli and Liang, Yan and Li, Yuanchun and Liu, Yunxin},
  journal={arXiv preprint arXiv:2508.05498},
  year={2025}
}
